\documentclass{article}

\PassOptionsToPackage{table}{xcolor}
\usepackage[preprint]{neurips_2026}

\usepackage[utf8]{inputenc}
\usepackage[T1]{fontenc}
\usepackage{hyperref}
\usepackage{url}
\usepackage{booktabs}
\usepackage{amsfonts}
\usepackage{nicefrac}
\usepackage{microtype}
\usepackage{xcolor}

\usepackage{graphicx}
\usepackage{amsmath}
\usepackage{amssymb}
\usepackage{amsthm}
\usepackage{multirow}
\usepackage{tabularx}
\usepackage{caption}
\usepackage{subcaption}
\usepackage{colortbl}
\usepackage{xfp}
\usepackage{siunitx}
\usepackage{fvextra}
\usepackage{float}
\usepackage{changepage}
\newlength{\figover}
\newcolumntype{Z}{>{\raggedright\arraybackslash}X}

\newcommand{\ysol}{y^{\star}}               
\newcommand{\yalt}{y'}                        
\newcommand{\ywrong}{y^{-}}                  
\newcommand{\yother}{\tilde{y}}              
\newcommand{\sstud}{s^{\mathrm{S}}}          
\newcommand{\steach}{s^{\mathrm{T}}}         
\newcommand{\pibias}{\mathrm{PS}}            
\newcommand{\JSD}{\mathrm{JSD}}             
\newcommand{\KL}{\mathrm{KL}}

\newcommand{\ph}[1]{\textcolor{red}{[#1]}}

\definecolor{chgpos}{RGB}{0,140,0}
\definecolor{chgneg}{RGB}{200,0,0}
\newcommand{\sdchgp}[3]{%
  \edef\tmpdiff{\fpeval{#3-#2}}%
  \ifdim\tmpdiff pt>0pt
    \textcolor{chgpos}{\num[print-implicit-plus,round-precision=#1]{\tmpdiff}}%
  \else\ifdim\tmpdiff pt<0pt
    \textcolor{chgneg}{\num[round-precision=#1]{\tmpdiff}}%
  \else
    \num[round-precision=#1]{0}%
  \fi\fi}
\newcommand{\sdchg}[2]{\sdchgp{1}{#1}{#2}}
\newcommand{\sdreset}{\gdef\sumTB{0}\gdef\sumTS{0}\gdef\cntT{0}\gdef\sumIB{0}\gdef\sumIS{0}\gdef\cntI{0}}
\NewDocumentCommand{\sdT}{m o m o}{%
  \xdef\sumTB{\fpeval{\sumTB+#1}}\xdef\sumTS{\fpeval{\sumTS+#3}}\xdef\cntT{\fpeval{\cntT+1}}%
  \IfNoValueTF{#2}{#1}{#1\,$\pm$\,#2} & \IfNoValueTF{#4}{#3}{#3\,$\pm$\,#4} & \sdchg{#1}{#3}}
\NewDocumentCommand{\sdI}{m o m o}{%
  \xdef\sumIB{\fpeval{\sumIB+#1}}\xdef\sumIS{\fpeval{\sumIS+#3}}\xdef\cntI{\fpeval{\cntI+1}}%
  \IfNoValueTF{#2}{#1}{#1\,$\pm$\,#2} & \IfNoValueTF{#4}{#3}{#3\,$\pm$\,#4} & \sdchg{#1}{#3}}

\newcommand{\sdavgT}{\xdef\abase{\fpeval{round(\sumTB/\cntT,2)}}\xdef\asd{\fpeval{round(\sumTS/\cntT,2)}}\num[round-precision=2]{\abase} & \num[round-precision=2]{\asd} & \sdchgp{2}{\abase}{\asd}}
\newcommand{\sdavgI}{%
  \ifnum\cntI=0
    \gdef\aIb{\ph{--}}\gdef\aIs{\ph{--}}\gdef\aId{\ph{--}}%
  \else
    \xdef\abasei{\fpeval{round(\sumIB/\cntI,2)}}\xdef\asdi{\fpeval{round(\sumIS/\cntI,2)}}%
    \gdef\aIb{\num[round-precision=2]{\abasei}}\gdef\aIs{\num[round-precision=2]{\asdi}}\gdef\aId{\sdchgp{2}{\abasei}{\asdi}}%
  \fi
  \aIb & \aIs & \aId}

\newcommand{\sdgreset}[1]{%
  \expandafter\gdef\csname sdB@#1\endcsname{0}%
  \expandafter\gdef\csname sdS@#1\endcsname{0}%
  \expandafter\gdef\csname sdN@#1\endcsname{0}}
\NewDocumentCommand{\sdg}{m m o m o}{%
  \expandafter\xdef\csname sdB@#1\endcsname{\fpeval{\csname sdB@#1\endcsname+#2}}%
  \expandafter\xdef\csname sdS@#1\endcsname{\fpeval{\csname sdS@#1\endcsname+#4}}%
  \expandafter\xdef\csname sdN@#1\endcsname{\fpeval{\csname sdN@#1\endcsname+1}}%
  \IfNoValueTF{#3}{#2}{#2\,$\pm$\,#3} & \IfNoValueTF{#5}{#4}{#4\,$\pm$\,#5} & \sdchg{#2}{#4}}

\newcommand{\sdgavg}[1]{%
  \edef\sdgN{\csname sdN@#1\endcsname}%
  \ifnum\sdgN=0
    \gdef\sdgAb{\ph{--}}\gdef\sdgAs{\ph{--}}\gdef\sdgAd{\ph{--}}%
  \else
    \xdef\sdgmB{\fpeval{round(\csname sdB@#1\endcsname/\sdgN,2)}}%
    \xdef\sdgmS{\fpeval{round(\csname sdS@#1\endcsname/\sdgN,2)}}%
    \gdef\sdgAb{\num[round-precision=2]{\sdgmB}}\gdef\sdgAs{\num[round-precision=2]{\sdgmS}}\gdef\sdgAd{\sdchgp{2}{\sdgmB}{\sdgmS}}%
  \fi
  \sdgAb & \sdgAs & \sdgAd}

\NewDocumentCommand{\sdb}{m m o}{%
  \gdef\sdcurbase{#2}%
  \expandafter\xdef\csname sdB@#1\endcsname{\fpeval{\csname sdB@#1\endcsname+#2}}%
  \expandafter\xdef\csname sdN@#1\endcsname{\fpeval{\csname sdN@#1\endcsname+1}}%
  \IfNoValueTF{#3}{#2}{#2\,$\pm$\,#3}}
\NewDocumentCommand{\sds}{m m o}{%
  \expandafter\xdef\csname sdS@#1\endcsname{\fpeval{\csname sdS@#1\endcsname+#2}}%
  \expandafter\xdef\csname sdN@#1\endcsname{\fpeval{\csname sdN@#1\endcsname+1}}%
  \IfNoValueTF{#3}{#2}{#2\,$\pm$\,#3} & \sdchg{\sdcurbase}{#2}}

\newcommand{\sdbavg}[1]{%
  \edef\sdbN{\csname sdN@#1\endcsname}%
  \ifnum\sdbN=0 \gdef\sdbA{\ph{--}}\else
    \xdef\sdbmB{\fpeval{round(\csname sdB@#1\endcsname/\sdbN,2)}}\gdef\sdbA{\num[round-precision=2]{\sdbmB}}\fi
  \sdbA}
\newcommand{\sdsavg}[2]{%
  \edef\sdsN{\csname sdN@#1\endcsname}%
  \edef\sdsNb{\csname sdN@#2\endcsname}%
  \ifnum\sdsN=0
    \gdef\sdsAs{\ph{--}}\gdef\sdsAd{\ph{--}}%
  \else
    \xdef\sdsmS{\fpeval{round(\csname sdS@#1\endcsname/\sdsN,2)}}%
    \gdef\sdsAs{\num[round-precision=2]{\sdsmS}}%
    \ifnum\sdsNb=0
      \gdef\sdsAd{\ph{--}}%
    \else
      \xdef\sdsmB{\fpeval{round(\csname sdB@#2\endcsname/\sdsNb,2)}}%
      \gdef\sdsAd{\sdchgp{2}{\sdsmB}{\sdsmS}}%
    \fi
  \fi
  \sdsAs & \sdsAd}

\NewDocumentCommand{\sdbq}{m m o}{%
  \gdef\sdcurbase{#2}%
  \expandafter\xdef\csname sdB@#1\endcsname{\fpeval{\csname sdB@#1\endcsname+#2}}%
  \expandafter\xdef\csname sdN@#1\endcsname{\fpeval{\csname sdN@#1\endcsname+1}}}
\NewDocumentCommand{\sdgs}{m m o m o}{%
  \expandafter\xdef\csname sdB@#1\endcsname{\fpeval{\csname sdB@#1\endcsname+#2}}%
  \expandafter\xdef\csname sdS@#1\endcsname{\fpeval{\csname sdS@#1\endcsname+#4}}%
  \expandafter\xdef\csname sdN@#1\endcsname{\fpeval{\csname sdN@#1\endcsname+1}}%
  \IfNoValueTF{#5}{#4}{#4\,$\pm$\,#5} & \sdchg{#2}{#4}}
\newcommand{\sdgavgs}[1]{%
  \edef\sdgN{\csname sdN@#1\endcsname}%
  \ifnum\sdgN=0
    \gdef\sdgAs{\ph{--}}\gdef\sdgAd{\ph{--}}%
  \else
    \xdef\sdgmB{\fpeval{round(\csname sdB@#1\endcsname/\sdgN,2)}}%
    \xdef\sdgmS{\fpeval{round(\csname sdS@#1\endcsname/\sdgN,2)}}%
    \gdef\sdgAs{\num[round-precision=2]{\sdgmS}}\gdef\sdgAd{\sdchgp{2}{\sdgmB}{\sdgmS}}%
  \fi
  \sdgAs & \sdgAd}

\definecolor{thinkbg}{RGB}{228,240,255}
\definecolor{instrbg}{RGB}{232,247,235}
\definecolor{avgbg}{RGB}{245,245,245}

\fvset{fontsize=\small,breaklines=true,breakanywhere=false,%
  breaksymbolleft={},breaksymbolright={},breakautoindent=true,%
  breakindent=1.5em,frame=single,framesep=6pt,rulecolor=\color{black!35},%
  numbers=none}

\newcounter{promptlisting}
\newcommand{\listingcaption}[1]{%
  \vspace{-2pt}%
  {\centering\small Listing~\refstepcounter{promptlisting}\thepromptlisting: #1\par}%
  \vspace{6pt}}

\newenvironment{plst}{\par\addvspace{\medskipamount}\noindent\begin{minipage}{\columnwidth}}%
  {\end{minipage}\par\addvspace{\medskipamount}}

\title{Privileged, but Biased: \\ How PI-Conditioned Teachers Break Self-Distillation}

\author{%
  \bfseries
  Sarthak Harne\hspace{0.8em}
  Chinmay Karkar\hspace{0.8em}
  Yash Pandya\hspace{0.8em}
  Ahmed Awadallah\hspace{0.8em}
  Akshay Nambi\thanks{Corresponding author: \texttt{akshayn@microsoft.com}} \\[2pt]
  Microsoft Research \\
}

\begin{document}
\maketitle
\vspace{-5pt}
\begin{abstract}
Self-distillation (SD) has emerged as a compute-efficient alternative to reinforcement learning with verifiable rewards: a self-teacher, conditioned on privileged information (PI) about the answer such as a reference solution, supplies dense per-token supervision to a student that never sees it. Reported gains, however, come almost exclusively from narrow, low-difficulty settings, leaving open a basic question: as a lone objective, with no reward term, does SD teach anything? We reproduce SDPO's reported gains in its easy setting, then apply the identical setup to difficult tasks and find that it does not. Across question answering, mathematics, coding, and multi-turn agentic tool use, across reasoning modes, model sizes, and forms of PI, and under both the SDPO and OPSD recipes, the per-token loss falls steadily while validation accuracy does not improve and typically degrades. We explain this failure through a single causal chain from the loss to the model it produces. The chain begins with PI bias: having seen one particular reference solution, the teacher's per-token target is pulled toward that trajectory rather than toward correctness in general, an effect we quantify with a PI Bias Score. Trained to match this target everywhere, the student's objective becomes nearly blind to whether a rollout is correct, and the loss it assigns falls mostly on low-information tokens like stopwords, punctuation, uncertainty markers, rather than those that determine the answer; within correct rollouts the exploratory tokens incur the highest divergence, so it penalizes the hesitation that reasoning requires. The result is a flatter, less decisive student that is no better at reasoning: as a lone objective, SD optimizes a signal decoupled from task success.
\end{abstract}
\vspace{-10pt}
\section{Introduction}
\label{sec:intro}
Reinforcement learning with verifiable rewards (RLVR) is a standard recipe for post-training large language models to reason \citep{grpo2024,rlvr2024}, but its supervision is coarse, sparse, and expensive: one scalar per rollout gives every token the same credit, many prompts yield no correct rollout, and each update needs a verifier and many on-policy samples. This motivates a denser signal supplied \textit{during} the rollout. Distillation \citep{distillm2024,OPD2024} supplies one but needs a separate, larger teacher; SD removes that requirement by building the teacher from the model being trained, conditioning it on privileged information (PI) the student is not given, such as a reference solution, a hint, or execution feedback \citep{sdft}. Conditioned on the PI the model is a much stronger next-token predictor of a correct trajectory, so matching student to teacher along the student's rollout turns a single privileged example into a dense per-token signal, with no separate teacher and no reward. SDPO \citep{sdpo2025} and OPSD \citep{opsd2025} use SD in RL post-training and report matching or beating GRPO at lower sample and compute cost, but on comparatively easy tasks such as short multiple-choice knowledge questions. \textbf{On difficult tasks, does SD as a lone objective, with no reward term, optimize task correctness, or a proxy that merely correlates with it easy tasks?}

We find that it does not optimize task correctness. We reproduce the reported SDPO behavior in its easy setting, then apply the identical setup to difficult tasks under four varied domains: general QA (MMLU-Pro), mathematics (DAPO-Math), coding (CodeForces), and multi-turn agentic tool use. We observe the opposite outcome: the per-token loss decreases steadily while validation accuracy does not improve and typically degrades, by up to $3.51$ points on average in the agentic domain and $7.0$ on individual benchmarks, under both recipes (Figure~\ref{fig:curves}, Section~\ref{sec:mainresults}). Since code, teacher, and clipping are unchanged, the difference lies in the objective, and not in the specific recipe.

The cause is two design choices the methods share: \textbf{PI-conditioning and per-token density jointly decouple the loss from correctness.} A teacher that has read one reference solution defines at each position a distribution shaped by that solution's surface form, not by correctness, and the dense divergence targets every token, including the many that do not decide the answer. Our central contribution makes this concrete as a \textbf{causal chain}: PI bias $\rightarrow$ a loss blind to correctness $\rightarrow$ effort spent on uninformative tokens $\rightarrow$ exploration penalized $\rightarrow$ a flattened student, every link measured on the same Qwen3-8B runs (Section~\ref{sec:chain}, Figures~\ref{fig:pibias} and~\ref{fig:chain}).

\textbf{(1) The teacher encodes one solution, not correctness.} Our \textit{PI Bias Score}, a teacher-minus-student log-probability ratio needing only forward passes, reaches $0.52$ on the in-context solution against ${\le}0.02$ on a \emph{different correct} solution to the same problem. This shows that the teacher barely separates from a solution to an unrelated problem. What the privileged information transfers is a specific trajectory instead of correctness.

\textbf{(2) A target built from one trajectory says nothing about the trajectory correctness.} Split by verifier outcome, per-token loss (${\approx}3.5\times10^{-4}$) and KL (${\approx}0.03$) overlap for correct and incorrect rollouts at every step. When they differ, it runs backwards, the teacher--student gap settles at $0.7$ on correct rollouts against $0.45$ on incorrect ones. This indicates that the learning pressure is highest where the student is already right, so the loss falls without accuracy following.

\textbf{(3) Density spreads that signal over tokens that do not decide the answer.} Stopwords, uncertainty markers, punctuation and whitespace absorb $55.4\%$ of the per-token loss, while the content words, numbers and math symbols that fix the answer absorb comparatively little.

\textbf{(4) Where the loss is high, it penalizes exploration.} Within \emph{correct} rollouts, positions off the reference path carry KL out to ${\approx}0.31$ against ${\approx}0.08$ on-path, a factor of four. Having read the solution, the teacher cannot tell a productive detour from a mistake, so the search that reasoning requires is penalized shortening the model responses.

\textbf{(5) The student flattens rather than sharpens.} The teacher--student gap closes from ${\approx}3.1$ to ${\approx}0$, but its $\pm1$ standard-deviation band does not contract and student entropy rises. The endpoint is a flatter, earlier-committing policy that solves no more problems. This becomes the training signature of Figure~\ref{fig:curves}.


Prior work saw fragments of this from opposite ends: RLSD \citep{rlsd2025} derives a conditional mutual-information \textit{leakage} bound but stops at theory, while \citet{reasoning_degrades2025} document degraded reasoning but leave its origin open; we show these are two ends of one mechanism, on difficult tasks where the method is actually used. 

\textbf{Our contributions are:} \textbf{(1)} evidence that SD, effective on easy tasks, fails as a lone objective on difficult reasoning, math, coding, and agentic tasks, holding across response length, task difficulty, model scale, reasoning mode, PI form, and both recipes (Sections~\ref{sec:signature},~\ref{sec:further}); \textbf{(2)} a causal chain of token- and distribution-level measurements explaining the failure (Section~\ref{sec:chain}); and \textbf{(3)} the \textit{PI Bias Score}, which measures how strongly a PI-conditioned teacher favors the in-context solution over other correct ones from forward passes alone. These reframe SD as an underspecified family of objectives rather than a method that does not work: the density that motivates it is intact; instead the target fails to encode correctness.

\section{Background and Related Work}
\label{sec:background}
SD sits between two signals it tries to combine: RLVR gives a coarse trajectory-level signal, while distillation gives a dense per-token signal but needs an external teacher. RLVR trains $\pi_\theta$ to maximize a binary reward $R(x,y)$ checking whether response $y$ to prompt $x$ is correct \citep{rlvr2024}, and GRPO \citep{grpo2024} estimates a group-relative advantage from $G$ rollouts per prompt with a PPO-style clipped update; every token receives the same advantage, and it vanishes when all rollouts in a group share a reward.

\textbf{Self-distillation: SDPO and OPSD.} SD replaces the reward with PI, side information about the answer available at training but not at test time; we use ``PI'' as an umbrella term for the feedback $f$ of SDPO \citep{sdpo2025}, the reference solution $\ysol$ of OPSD \citep{opsd2025} and RLSD \citep{rlsd2025}, and the conditioning context $c$ of \citet{reasoning_degrades2025}. Training minimizes a stop-gradient per-token divergence between teacher and student along a student rollout (Eq.~\eqref{eq:sd}), so no external model or reward is required. SDPO uses a per-token KL to a teacher conditioned on $f$, ranging from a prior successful rollout to environment output such as failed unit tests and stabilized by an EMA or trust-region teacher; it matches or exceeds GRPO at several-fold lower compute on SciKnowEval, ToolAlpaca and LiveCodeBench. OPSD uses a forward $\KL(p_T\|p_S)$ with the teacher conditioned on $\ysol$ and fixed to the initial policy, observes that a few stylistic tokens carry far higher divergence than answer-determining ones, casts the update as a dense token-level rather than sequence-level policy gradient, and reports gains over GRPO and off-policy distillation on AIME \citep{aime2024, aime2025} and HMMT \citep{hmmt2024}.

\textbf{Critiques and augmentations.} RLSD \citep{rlsd2025} proves the objective decomposes as $\mathcal{L}_{\mathrm{OPSD}}=\mathcal{L}^{*}+I(Y_t;R\mid X,Y_{<t})$, an irreducible conditional mutual information: the teacher fits a single PI-conditioned trace rather than the marginal over valid solutions, so a teacher favoring one phrasing trains the student to adopt it; RLSD's fix uses the teacher only to reweight a verifier-driven advantage. \citet{reasoning_degrades2025} report that richer teacher conditioning makes reasoning more concise and confident, speeding in-domain optimization but degrading out-of-distribution and on harder problems by up to $\sim$40\% across the Qwen3 and DeepSeek-R1-Distill families. A further line keeps a verifiable reward primary and enters SD as a small-weight auxiliary regularizer \citep{skillsd2025,sdar2025,opid2026} in multi-turn and agentic settings. Across all of this work SD as a \emph{lone} objective on difficult tasks is never tested; that is the gap we take up, tying the leakage and uncertainty-suppression accounts to one set of measured training runs.

\begin{figure}[tb]
\begin{adjustwidth}{-\figover}{-\figover}
\centering
\begin{subfigure}{0.2\linewidth}
  \centering
  \includegraphics[width=\linewidth]{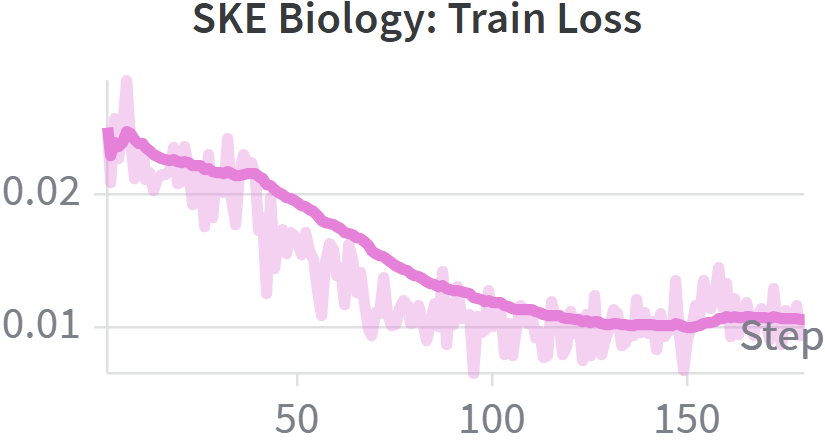}
  \caption{Biology: loss}
  \label{fig:repro-bio-loss}
\end{subfigure}
\begin{subfigure}{0.2\linewidth}
  \centering
  \includegraphics[width=\linewidth]{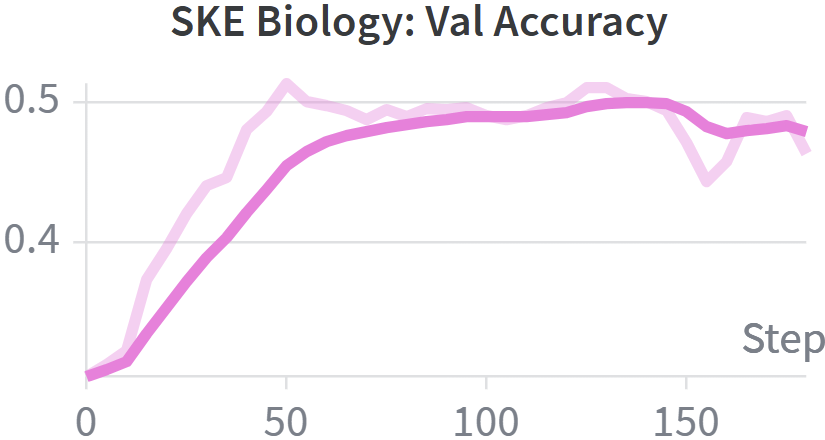}
  \caption{Biology: accuracy}
  \label{fig:repro-bio-acc}
\end{subfigure}
\begin{subfigure}{0.2\linewidth}
  \centering
  \includegraphics[width=\linewidth]{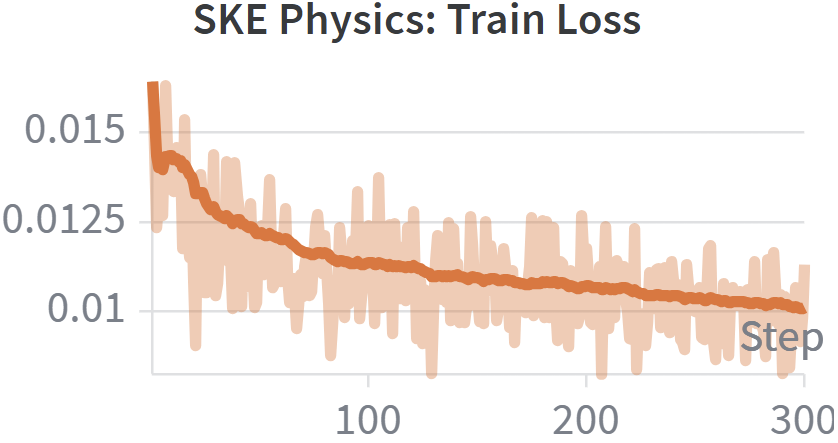}
  \caption{Physics: loss}
  \label{fig:repro-phys-loss}
\end{subfigure}
\begin{subfigure}{0.2\linewidth}
  \centering
  \includegraphics[width=\linewidth]{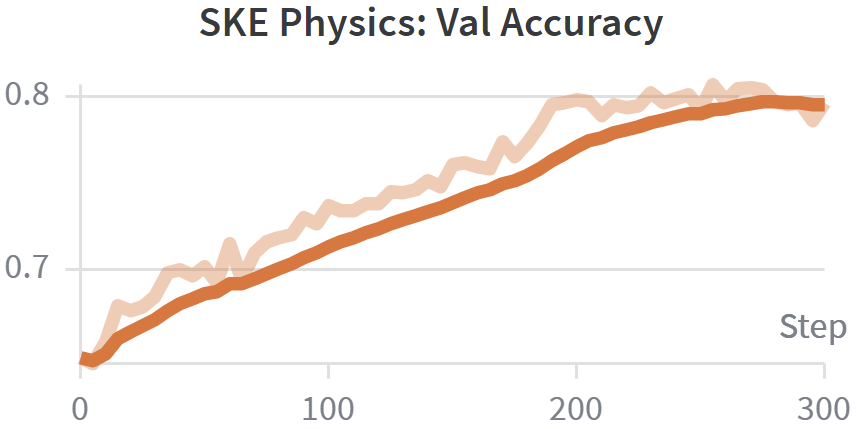}
  \caption{Physics: accuracy}
  \label{fig:repro-phys-acc}
\end{subfigure}
\end{adjustwidth}
\caption{\textbf{Reproducing SD in its original regime.}
Qwen3-8B on SciKnowEval with whole-solution PI. Loss decreases and validation
\emph{rises} on Biology and Physics, matching the reported SDPO trend.}
\label{fig:repro}
\end{figure}

\section{The SD Objective and Two Sources of Bias}
\label{sec:objective}

A policy $\pi_\theta$ generates $y=(y_1,\dots,y_T)$ for a prompt $x$ autoregressively. The student is conditioned on the prompt alone and the self-teacher is the same model additionally conditioned on the PI $r$: $p_S(\cdot\mid x,y_{<t})=\pi_\theta(\cdot\mid x,y_{<t})$ and $p_T(\cdot\mid x,y_{<t})=\pi_{\theta'}(\cdot\mid x,r,y_{<t})$, where $\theta'$ is the student's own parameters or a slow copy that tracks them. Conditioning on $r$ makes $p_T$ a stronger next-token predictor of a correct continuation; the student, which never sees $r$, is trained to match it along a rollout $\hat y\sim p_S$ from the student itself:
\begin{equation}
\resizebox{\linewidth}{!}{$\displaystyle
\mathcal{L}_{\mathrm{SD}}(\theta)=\mathbb{E}_{(x,r)}\,\mathbb{E}_{\hat y\sim \mathrm{sg}(p_S)}\!\left[\frac{1}{|\hat y|}\sum_{t=1}^{|\hat y|} D\!\big(\mathrm{sg}(p_T(\cdot\mid x,r,\hat y_{<t}))\,\big\|\,p_S(\cdot\mid x,\hat y_{<t})\big)\right],
$}
\label{eq:sd}
\end{equation}
where $\mathrm{sg}$ is stop-gradient, so gradients flow only through the student. SDPO and OPSD instantiate $D$ differently; we study Eq.~\eqref{eq:sd} as a \textbf{lone objective}, with no reward term. Two of its properties drive everything that follows.

\textbf{Property 1: a dense per-token target.} Unlike a trajectory-level reward, Eq.~\eqref{eq:sd} imposes a target at every position; both SDPO and OPSD analyze this as a dense token-level policy gradient (SDPO via its per-token KL, their Proposition~2.1, and OPSD by contrast with sequence-level self taught reasoning). Density is the design's advantage, but it also places punctuation, stopwords, and style-carrying discourse markers on equal footing with content-bearing tokens.

\textbf{Property 2: a PI-conditioned target.} With a whole reference solution $r=\ysol$ the target reflects one particular solution, in both the answer it reaches and how it is phrased and ordered; matching it token by token asks the student to reproduce that behavior everywhere. RLSD \citep{rlsd2025} formalizes the cost as an irreducible $I(Y_t;R\mid X,Y_{<t})$, measuring how much the target depends on the PI beyond what correctness requires. A dense loss aimed at one solution's surface form therefore spreads learning signal across all tokens rather than the decisions that make an answer correct. We explore this further link by link in Section~\ref{sec:chain}.

\section{Experimental Setup}
\label{sec:setup}
\paragraph{Domains and data.}
Prior work evaluates SD on short, knowledge-recall tasks. To test the objective where it is meant to be used, we assemble four harder domains: general QA uses MMLU-Pro \citep{mmlupro2024}, extending SDPO's SciKnowEval format to ten options and reasoning over recall; mathematics uses the de-duplicated DAPO-Math-17k \citep{dapo2025,dapomathprocessed2025}, whose spread of response lengths separates trace length from difficulty (Section~\ref{sec:lengthsplit}); coding uses CodeForces \citep{codeforcesdata2025}, demanding an algorithm rather than a short completion; and the agentic setting uses BFCL \citep{bfcl2024}, requiring state across turns and composed tool calls. We train on 2{,}000 examples for general QA, math, and coding, and on the BFCL multiturn split, evaluating every domain \textbf{in-domain} and on \textbf{held-out transfer benchmarks} from different sources, so a gain cannot come from fitting the training distribution. Transfer benchmarks are SciKnowEval \citep{sciknoweval2024} and GPQA-D \citep{gpqa2023}; AIME24 \citep{aime2024}, AIME25 \citep{aime2025}, Olympiad Bench \citep{olympiadbench2024}; MBPP+ \citep{mbpp2021}, HumanEval+ \citep{humaneval2021}, CodeElo \citep{codeelo2025}, LCBv6 \citep{lcb2025}; and BFCLv4 multiturn \citep{bfcl2024}. Data details are in Appendix~\ref{app:data}, prompts and templates in Appendix~\ref{app:prompts}.

\paragraph{Models and privileged information.}
Our primary model is Qwen3-8B \citep{qwen32025} in both its think (long chain-of-thought) and instruct variants, so every domain is studied under both reasoning modes; for a size comparison we additionally train Qwen3-32B on general QA. Our default PI is the whole reference solution $r=\ysol$: following \citet{sdpo2025} we sample multiple rollouts per query and take one random correct rollout. Section~\ref{sec:pichoice} additionally studies short hints (one to two sentences) and skills ($\approx$500-token structured guides of relevant techniques).

\paragraph{Objective, divergence, and training.}
We train with the lone objective of Eq.~\eqref{eq:sd}, no reward or verifier in the loss, instantiating $D$ as the symmetric Jensen--Shannon divergence, $\JSD(p\,\|\,q)=\tfrac12\KL(p\,\|\,m)+\tfrac12\KL(q\,\|\,m)$ with $m=\tfrac12(p+q)$, matching SDPO's implementation; the OPSD variant is reported in Section~\ref{sec:mainresults}. Two ingredients stabilize training on harder data, both motivated by the loss concentration of Section~\ref{sec:arg2}: aggressive per-token clipping of the divergence \citep{opsd2025}, preventing a few high-divergence stylistic tokens from dominating the gradient, and a slow EMA teacher $\theta'\leftarrow(1-\alpha)\theta'+\alpha\theta$ with $\alpha{=}0.001$ (vs.\ $0.01$ in SDPO and $0.05$ in \citet{reasoning_degrades2025}), which removed the late-training instabilities we first observed. Hyperparameters: JSD clipping $\tau{=}0.001$, learning rate $1\mathrm{e}{-5}$, batch size $32$, $3$ epochs (Appendix~\ref{app:hyper}). We report task accuracy (math, QA), fraction of test cases passed (coding), and the BFCL success metric (agentic), each averaged over 4 sampled val trajectories per task at each validation step, and track the per-token SD loss, validation accuracy, student entropy $\mathbb{H}[p_S(\cdot\mid x,\hat y_{<t})]$, and average response length, plus teacher--student KL and teacher perplexity for Section~\ref{sec:arg5}.

\paragraph{Reproducing SD in its original regime.}
Using whole-solution PI, we train Qwen3-8B with Eq.~\eqref{eq:sd} on SciKnowEval with SDPO's codebase \citep{sdpo2025} and reported hyperparameters, and recover the qualitative SDPO result: on Biology and Physics the loss decreases while validation accuracy rises and responses shorten (Figure~\ref{fig:repro}), confirming that implementation is faithful. We further add on to the SDPO codebase for all our experiments. Our code is available as part of the supplementary material.

\section{The Training Signature of Lone SD }
\label{sec:signature}
Applying the same implementation to the four domains produces a consistent training signature: the loss is optimized, and nothing that determines task success improves. Harder data also changes where the loss lives. Unclipped, the top 50\% of tokens consume 98.5\% of the loss on MMLU-Pro, mirroring the skew OPSD \citep{opsd2025} reports on math: a few high-divergence tokens dominate the gradient, so the average falls while little signal reaches content tokens. This is what the clipping and slow teacher of Section~\ref{sec:setup} address.

\subsection{Patterns observed during training}
\label{sec:diagnostics}

\begin{figure*}[t]
\begin{adjustwidth}{-\figover}{-\figover}
\centering
\begin{subfigure}{0.2\linewidth}\centering
  \includegraphics[width=\linewidth]{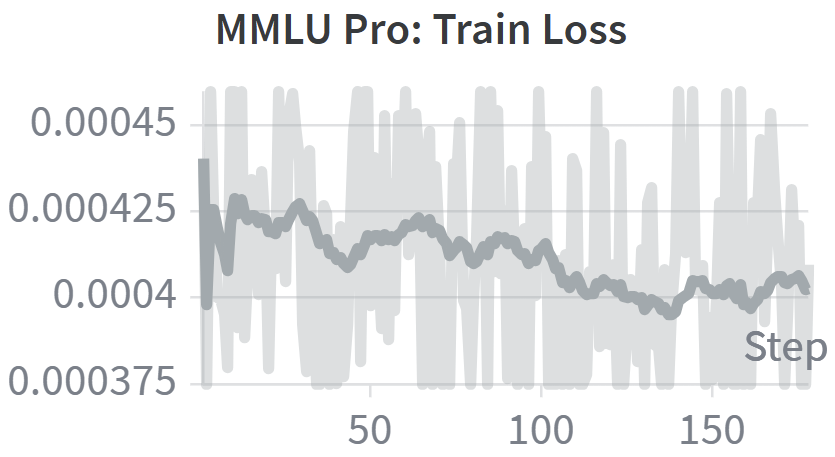}
\end{subfigure}
\begin{subfigure}{0.2\linewidth}\centering
  \includegraphics[width=\linewidth]{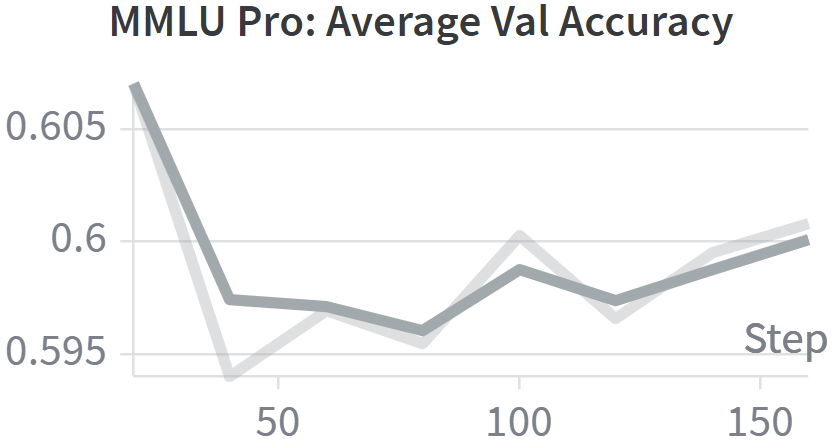}
\end{subfigure}
\begin{subfigure}{0.2\linewidth}\centering
  \includegraphics[width=\linewidth]{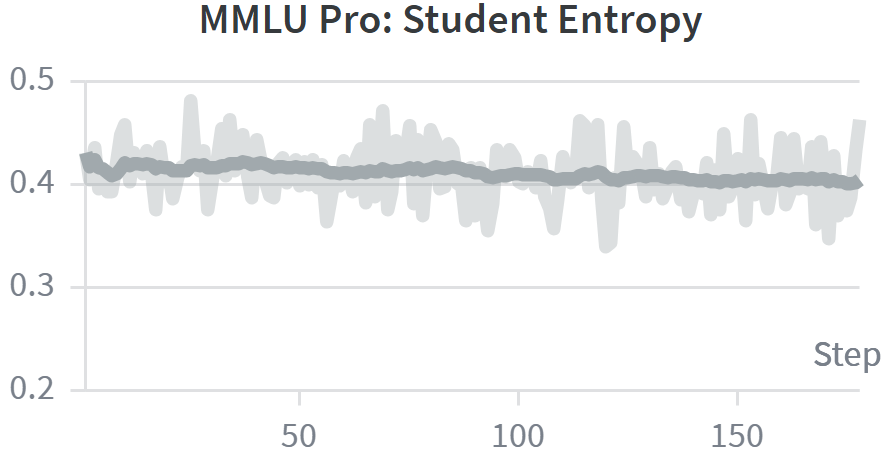}
\end{subfigure}
\begin{subfigure}{0.2\linewidth}\centering
  \includegraphics[width=\linewidth]{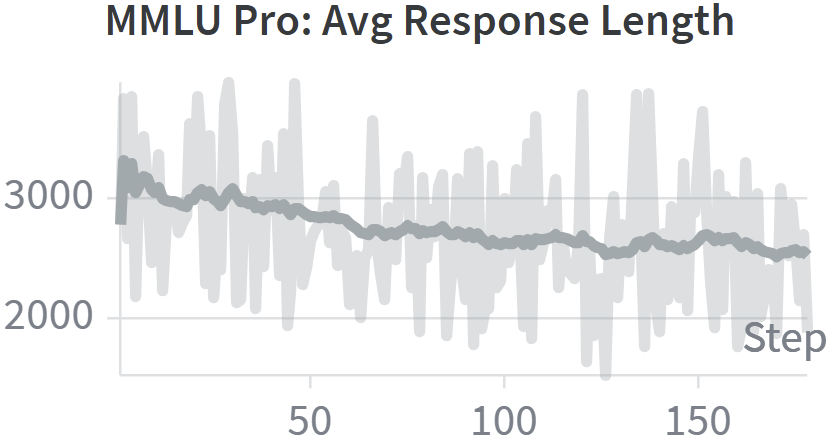}
\end{subfigure}

\vspace{0.8em}

\begin{subfigure}{0.2\linewidth}\centering
  \includegraphics[width=\linewidth]{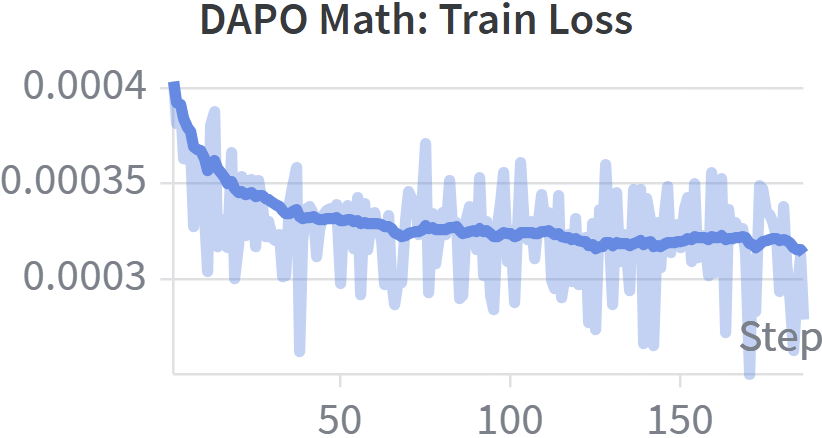}
\end{subfigure}
\begin{subfigure}{0.2\linewidth}\centering
  \includegraphics[width=\linewidth]{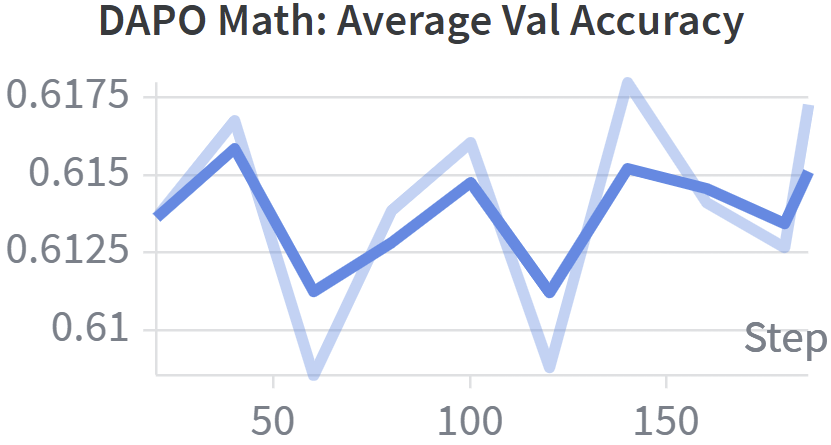}
\end{subfigure}
\begin{subfigure}{0.2\linewidth}\centering
  \includegraphics[width=\linewidth]{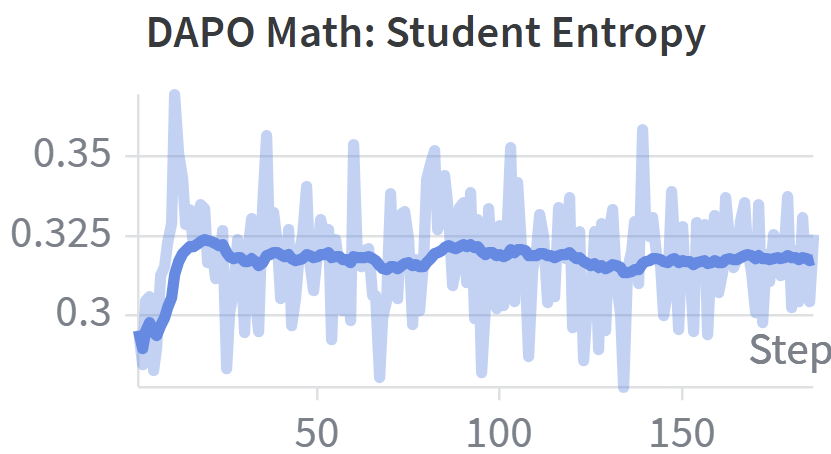}
\end{subfigure}
\begin{subfigure}{0.2\linewidth}\centering
  \includegraphics[width=\linewidth]{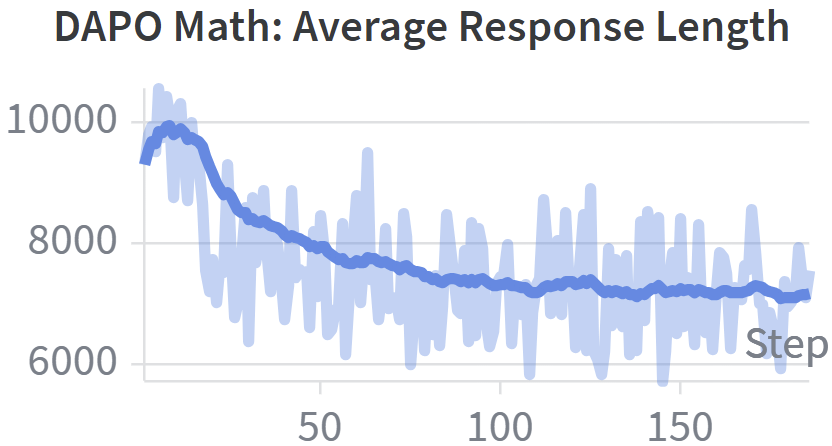}
\end{subfigure}

\vspace{0.8em}

\begin{subfigure}{0.2\linewidth}\centering
  \includegraphics[width=\linewidth]{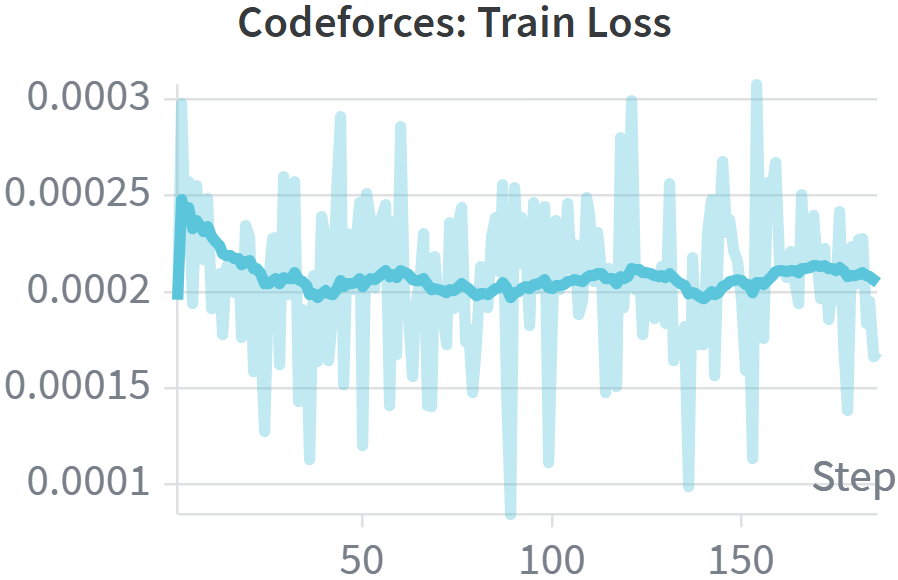}
\end{subfigure}
\begin{subfigure}{0.2\linewidth}\centering
  \includegraphics[width=\linewidth]{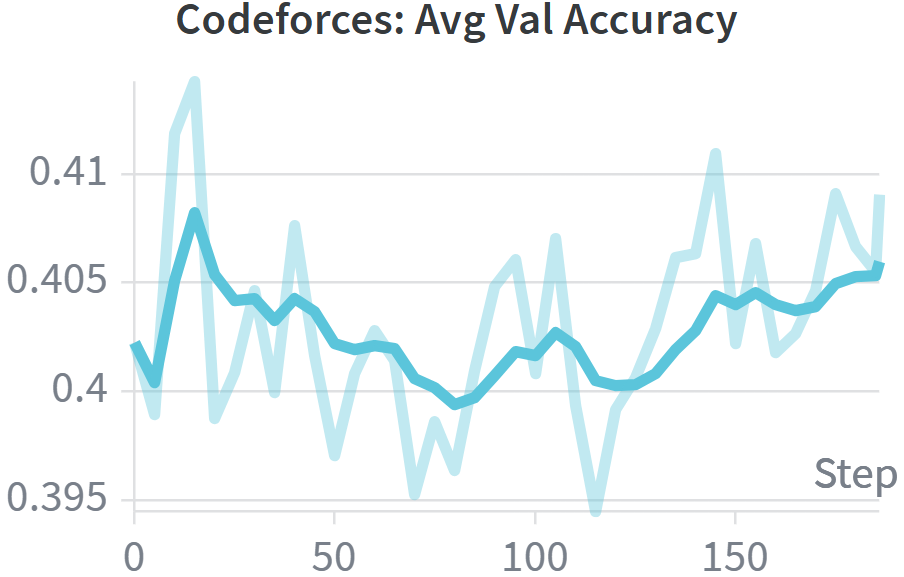}
\end{subfigure}
\begin{subfigure}{0.2\linewidth}\centering
  \includegraphics[width=\linewidth]{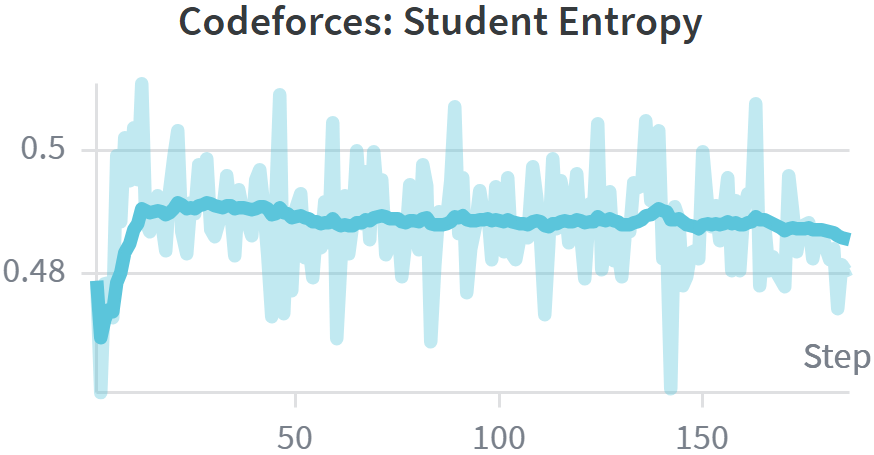}
\end{subfigure}
\begin{subfigure}{0.2\linewidth}\centering
  \includegraphics[width=\linewidth]{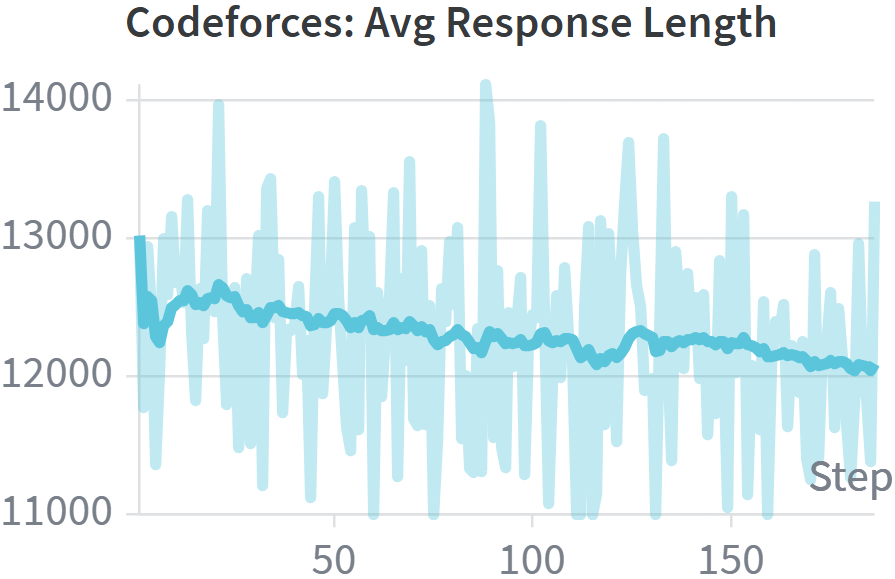}
\end{subfigure}

\vspace{0.8em}

\begin{subfigure}{0.2\linewidth}\centering
  \includegraphics[width=\linewidth]{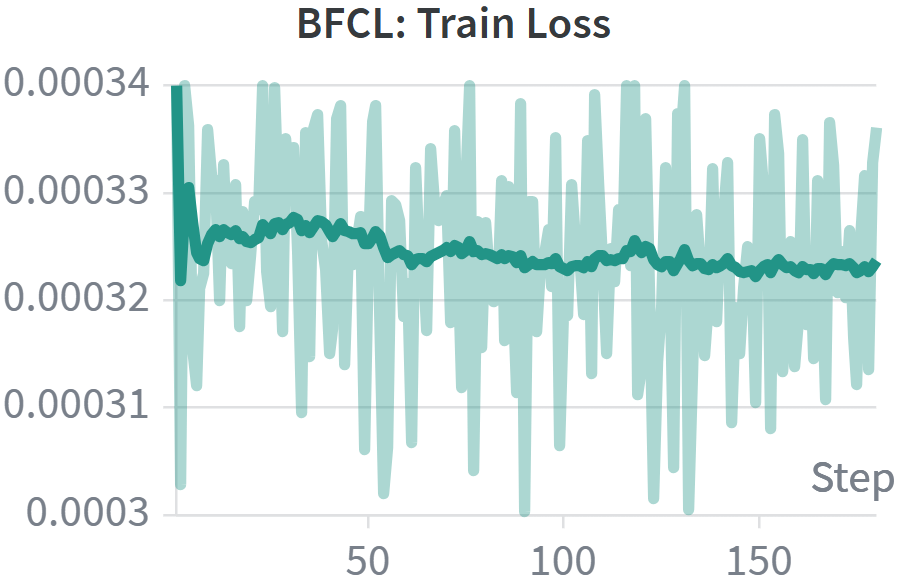}
\end{subfigure}
\begin{subfigure}{0.2\linewidth}\centering
  \includegraphics[width=\linewidth]{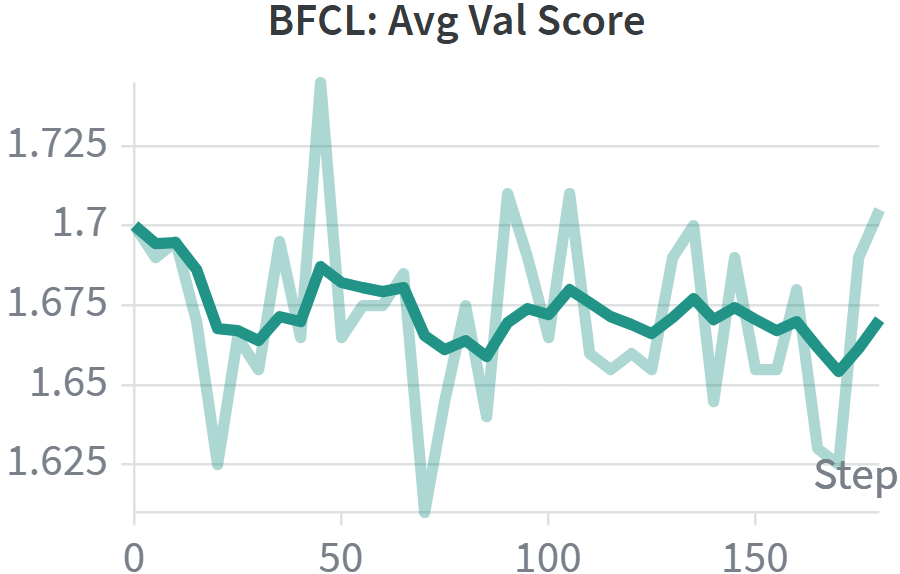}
\end{subfigure}
\begin{subfigure}{0.2\linewidth}\centering
  \includegraphics[width=\linewidth]{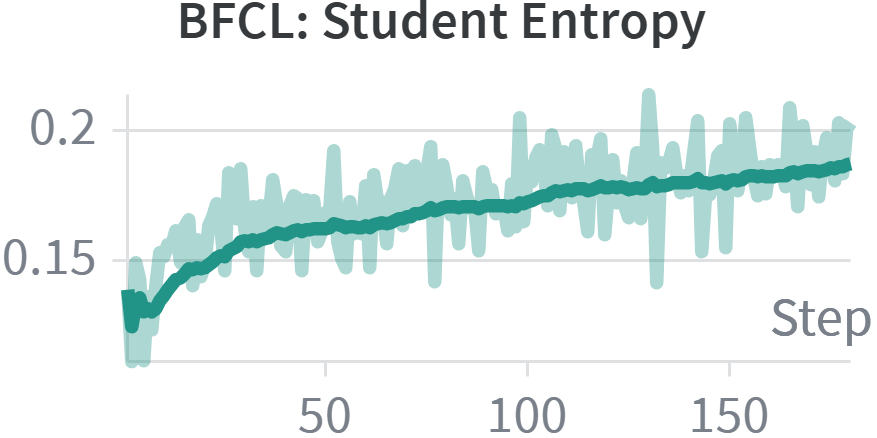}
\end{subfigure}
\begin{subfigure}{0.2\linewidth}\centering
  \includegraphics[width=\linewidth]{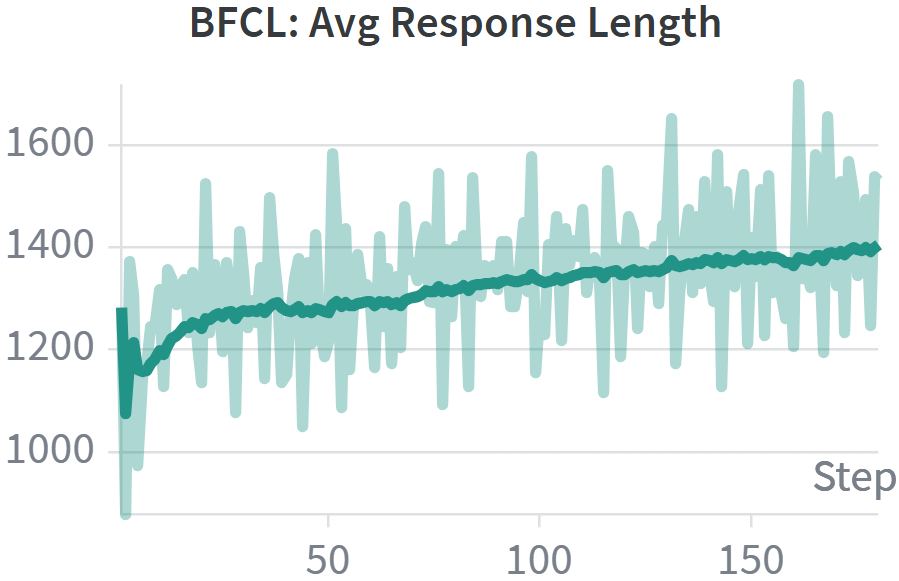}
\end{subfigure}
\end{adjustwidth}
\caption{Per-domain training dynamics for lone self-distillation on Qwen3-8B.
Rows, from top to bottom, correspond to general QA (MMLU-Pro), mathematics
(DAPO-Math), coding (CodeForces), and multi-turn agentic tool use (BFCL).
Columns, from left to right, report the per-token loss, the validation accuracy,
the student entropy, and the mean response length over training. Across all four
domains the loss decreases while the validation accuracy does not improve. For
BFCL the second column reports the validation score rather than the accuracy; the
score equals twice the accuracy.}
\label{fig:curves}
\end{figure*}

Figure~\ref{fig:curves} shows the training dynamics for Qwen3-8B, and the same pattern recurs across settings. \textbf{The loss is optimized:} the per-token JS divergence of Eq.~\eqref{eq:sd} decreases steadily, indicating closer matching to the teacher. \textbf{Validation accuracy does not follow:} it remains flat or degrades, so optimization is decoupled from task performance. \textbf{Student entropy rises:} $\mathbb{H}[p_S(\cdot\mid x,\hat y_{<t})]$ increases, contrary to the sharpening expected of an improving model (Section~\ref{sec:arg5}). \textbf{Response length falls:} while desirable for short-form generation, on reasoning tasks this is also consistent with reduced exploration before committing (Section~\ref{sec:arg4}). Any single trend admits a benign interpretation; together they indicate a systematic failure mode. The pattern is consistent across all four domains and both reasoning modes. The only exception is agentic instruct mode, where the small training set and short responses (300 tokens) yield a modest validation improvement. Because the domains differ substantially in response length, credit structure, and reference-solution format, the shared behavior is most naturally attributed to the objective rather than any particular dataset. Training curves for all domains are provided in Figure~\ref{fig:curves}.

Figure~\ref{fig:curves} shows the per-domain training dynamics for
self-distillation used as the sole training signal on Qwen3-8B. Each row
corresponds to one domain, and each column reports one diagnostic over training:
the per-token loss, the validation accuracy, the student entropy, and the mean
response length. Across all four domains the per-token loss decreases while the
validation accuracy does not improve, and the entropy and response length move
in the directions described in the main paper.

\definecolor{thinkbg}{RGB}{228,240,255}
\definecolor{instrbg}{RGB}{232,247,235}
\definecolor{avgbg}{RGB}{245,245,245}
\begin{table}[tb]
\centering
\caption{\textbf{SD as a lone objective does not improve validation performance across various domains.} Positive $\Delta$ in green, negative in red; averages over four independent rollouts.}
\label{tab:main}

\footnotesize
\setlength{\tabcolsep}{3pt}
\renewcommand{\arraystretch}{1.2}

\resizebox{\columnwidth}{!}{%
\begin{tabular}{llcccccc}
\toprule

&
&
\multicolumn{3}{>{\columncolor{thinkbg}}c}{\textbf{Think}} &
\multicolumn{3}{>{\columncolor{instrbg}}c}{\textbf{Instruct}}\\

\cmidrule(lr){3-5}
\cmidrule(lr){6-8}

\textbf{Domain} &
\textbf{Benchmark} &
\textbf{Base} &
\textbf{SD} &
$\mathbf{\Delta}$ &
\textbf{Base} &
\textbf{SD} &
$\mathbf{\Delta}$\\

\midrule

\noalign{\sdreset}
\rowcolor{gray!6}
\multirow{7}{*}{\shortstack[c]{General\\QA}}
& MMLU-Pro
& \sdT{70.2}[2.2]{69.8}[2.2]
& \sdI{69.5}[2.1]{67.5}[2.3]
\\

& GPQA-D
& \sdT{59.5}[3.0]{57.6}[3.2]
& \sdI{49.2}[2.7]{46.0}[2.8]
\\

& Bio (SKE)
& \sdT{38.5}[3.4]{36.7}[3.6]
& \sdI{33.5}[3.5]{32.3}[3.4]
\\

& Chem (SKE)
& \sdT{51.9}[2.0]{51.9}[2.0]
& \sdI{44.6}[1.8]{41.5}[2.0]
\\

& Mat. (SKE)
& \sdT{68.0}[3.2]{64.6}[3.1]
& \sdI{54.7}[3.1]{54.9}[3.4]
\\

& Phys. (SKE)
& \sdT{77.5}[2.8]{75.6}[3.0]
& \sdI{64.2}[3.2]{64.1}[3.3]
\\

\rowcolor{avgbg}
& \textbf{Average}
& \sdavgT
& \sdavgI
\\

\midrule

\noalign{\sdreset}
\rowcolor{gray!6}
\multirow{5}{*}{Math}
& DAPO
& \sdT{79.1}[2.0]{80.1}[2.0]
& \sdI{48.8}[2.2]{44.8}[2.4]
\\

& AIME24
& \sdT{67.5}[7.7]{66.0}[7.8]
& \sdI{24.2}[7.0]{23.6}[7.0]
\\

& AIME25
& \sdT{53.3}[7.4]{52.9}[7.4]
& \sdI{20.0}[6.2]{17.2}[6.1]
\\

& Olympiad
& \sdT{53.6}[1.9]{53.4}[1.8]
& \sdI{46.3}[1.7]{46.5}[1.7]
\\

\rowcolor{avgbg}
& \textbf{Average}
& \sdavgT
& \sdavgI
\\

\midrule

\noalign{\sdreset}
\rowcolor{gray!6}
\multirow{6}{*}{Coding}
& CodeForces
& \sdT{24.5}[1.6]{25.0}[1.8]
& \sdI{17.3}[1.3]{16.4}[1.3]
\\

& MBPP+
& \sdT{59.3}[2.3]{53.1}[2.1]
& \sdI{43.2}[2.0]{55.8}[2.3]
\\

& HumanEval+
& \sdT{80.5}[3.1]{80.4}[2.9]
& \sdI{77.4}[2.8]{73.6}[3.1]
\\

& CodeElo
& \sdT{23.8}[1.5]{25.1}[1.9]
& \sdI{13.8}[1.1]{12.1}[1.3]
\\

& LCBv6
& \sdT{50.0}[3.9]{53.0}[3.7]
& \sdI{52.3}[2.8]{50.8}[3.2]
\\

\rowcolor{avgbg}
& \textbf{Average}
& \sdavgT
& \sdavgI
\\

\midrule

\noalign{\sdreset}
\rowcolor{gray!6}
\multirow{6}{*}{Agentic}
& Base V3
& \sdT{86.0}[3.2]{83.2}[4.6]
& \sdI{68.0}[4.6]{72.5}[4.2]
\\

& Base V4
& \sdT{51.3}[2.9]{48.8}[2.9]
& \sdI{23.5}[2.8]{31.0}[3.2]
\\

& Miss Func
& \sdT{51.0}[3.0]{44.0}[2.9]
& \sdI{19.5}[2.4]{21.5}[2.7]
\\

& Miss Param
& \sdT{33.0}[2.9]{31.36}[2.7]
& \sdI{17.0}[2.3]{19.88}[2.7]
\\

& Long Ctx
& \sdT{35.5}[2.9]{31.88}[2.7]
& \sdI{16.5}[2.5]{23.0}[2.9]
\\

\rowcolor{avgbg}
& \textbf{Average}
& \sdavgT
& \sdavgI
\\

\bottomrule
\end{tabular}}
\end{table}

\subsection{End-task results}
\label{sec:mainresults}

\textbf{No improvement where reasoning is required.} Table~\ref{tab:main} evaluates each trained model against its own base, in-domain and on transfer. Across the reasoning-heavy domains, in both modes, the objective produces no improvement, and degradation is largest where reasoning demand is highest: the think variant loses $3.51$ points on average in the agentic domain, including $7$ on BFCL Multi-Turn Missing Function V4, and both modes lose $1.6$ across general QA. In-domain and transfer move together, consistent with the objective failing to install a capability rather than installing a narrow one.

\textbf{Positive averages appear only under the instruct variant.} Two domain averages come out positive, neither a gain. The instruct variant improves on the agentic BFCL benchmarks, but BFCL trains on a much smaller split (100 tasks) with short instruct rollouts (mean 300 tokens). In coding, the instruct gain on MBPP+ ($+12.6$) is offset by losses on HumanEval+ ($-3.8$), CodeElo ($-1.7$), LCBv6 ($-1.5$) and CodeForces ($-0.9$), so performance still declines averaged over benchmarks and modes. Both cases are confined to the instruct model, where shorter responses limit the reach of the causal chain; under the think, flat or declining is the norm.

\textbf{The result is not specific to SDPO.} We repeat the evaluation under the OPSD recipe, a single rollout per step, fixed correct answers, and an initial-policy teacher. Averages change by $-2.0$ and $-0.7$ points on general QA and $-4.3$ and $-0.4$ on math (think and instruct), with 18 of 20 benchmark--mode cells non-positive and the same signature throughout (full table in Table~\ref{tab:opsd}). Since the recipes differ in divergence direction, teacher schedule, rollout count, and clipping, the behavior tracks the shared structure of Eq.~\eqref{eq:sd} rather than one method's implementation choices.

\begin{table}[t]
\centering
\caption{\textbf{The same non-improvement pattern holds under the OPSD recipe.}
Results on General QA and Mathematics using the OPSD training recipe.
Positive $\Delta$ values are shown in green and negative values in red.
All values are averaged over four independent rollouts.}
\label{tab:opsd}

\footnotesize
\setlength{\tabcolsep}{2.5pt}
\renewcommand{\arraystretch}{1.0}

\resizebox{\columnwidth}{!}{%
\begin{tabular}{llcccccc}
\toprule
&
&
\multicolumn{3}{>{\columncolor{thinkbg}}c}{\textbf{Think}} &
\multicolumn{3}{>{\columncolor{instrbg}}c}{\textbf{Instruct}}\\
\cmidrule(lr){3-5}
\cmidrule(lr){6-8}
\textbf{Domain} &
\textbf{Benchmark} &
\textbf{Base} &
\textbf{SD} &
$\mathbf{\Delta}$ &
\textbf{Base} &
\textbf{SD} &
$\mathbf{\Delta}$\\
\midrule
\noalign{\sdreset}
\rowcolor{gray!6}
\multirow{7}{*}{\shortstack[l]{General\\QA}}
& MMLU-Pro
& \sdT{70.2}[2.2]{70.1}[2.2]
& \sdI{69.5}[2.1]{69.1}[2.3]
\\
&
GPQA-D
& \sdT{59.5}[3.0]{57.1}[3.0]
& \sdI{49.2}[2.7]{46.6}[2.8]
\\
&
Bio (SKE)
& \sdT{38.5}[3.4]{35.1}[3.5]
& \sdI{33.5}[3.5]{32.6}[3.4]
\\
&
Chem (SKE)
& \sdT{51.9}[2.0]{51.2}[2.0]
& \sdI{44.6}[1.8]{43.0}[2.0]
\\
&
Mat. (SKE)
& \sdT{68.0}[3.2]{64.2}[3.1]
& \sdI{54.7}[3.1]{56.6}[3.4]
\\
&
Phys. (SKE)
& \sdT{77.5}[2.8]{76.1}[3.0]
& \sdI{64.2}[3.2]{63.9}[3.3]
\\
\rowcolor{avgbg}
&
\textbf{Average}
&
\sdavgT
&
\sdavgI
\\
\midrule
\noalign{\sdreset}
\rowcolor{gray!6}
\multirow{5}{*}{\shortstack[l]{Math}}
& DAPO
& \sdT{79.1}[2.0]{76.0}[1.9]
& \sdI{48.8}[2.2]{46.8}[2.4]
\\
&
AIME24
& \sdT{67.5}[7.7]{62.2}[7.3]
& \sdI{24.2}[7.0]{23.7}[7.0]
\\
&
AIME25
& \sdT{53.3}[7.4]{48.2}[8.2]
& \sdI{20.0}[6.2]{18.6}[6.1]
\\
&
Olympiad
& \sdT{53.6}[1.9]{50.0}[1.8]
& \sdI{46.3}[1.7]{48.4}[1.9]
\\
\rowcolor{avgbg}
&
\textbf{Average}
&
\sdavgT
&
\sdavgI
\\
\bottomrule
\end{tabular}}
\end{table}

\newcommand{\sddelta}[1]{%
  \ifnum\fpeval{#1 > 0}=1
    \textbf{\textcolor{chgpos}{$+#1$}}%
  \else
    \ifnum\fpeval{#1 < 0}=1
      \textbf{\textcolor{chgneg}{$#1$}}%
    \else
      \textbf{\textcolor{gray}{0.0}}%
    \fi
  \fi
}
\vspace{-5pt}
\paragraph{What the signature rules out.}
\label{sec:rulesout}
Three explanations for falling loss and flat accuracy are ruled out above. \textbf{Optimization failure:} the loss decreases smoothly under the stabilization of Section~\ref{sec:setup}. \textbf{Implementation error:} the same code, teacher, and clipping reproduce the reported gains on SciKnowEval (Figure~\ref{fig:repro}); what changes is task difficulty. What remains is the objective itself.
\section{A Mechanistic Account: From Privileged Information to a Misguided Student}
\label{sec:chain}

We now measure each link of the predicted chain in causal order, beginning with the one we can intervene on: the bias is a property of the PI, and changing the PI's specificity moves it as predicted, so the chain does not rest on correlation alone. All measurements use Qwen3-8B with whole-solution PI; a rollout is \textit{correct} if the student solved the problem.

\subsection{PI bias: the teacher targets one solution, not correctness}
\label{sec:arg3}

\paragraph{An instrument for the bias.}
To measure what the teacher moves the student toward, fix a rollout prefix $(x,y_{<t})$ and a target sequence $w=(w_1,\dots,w_K)$. The student and teacher scores are the average log-probability each assigns to $w$, and the \textbf{PI Bias Score} is their difference:
\begin{equation}
\pibias_t(w)
=
\steach_t(w)-\sstud_t(w)
=
\frac{1}{K}\sum_{k=1}^{K}
\log
\frac{p_T(w_k\mid x,\ysol,y_{<t})}
{p_S(w_k\mid x,y_{<t})}.
\label{eq:pibias}
\end{equation}
This discounts what the student already assigns to $w$, reporting only how much the PI moves mass toward it (Appendix~\ref{app:pibias}).

\begin{figure}[b]
\centering
\includegraphics[width=\linewidth]{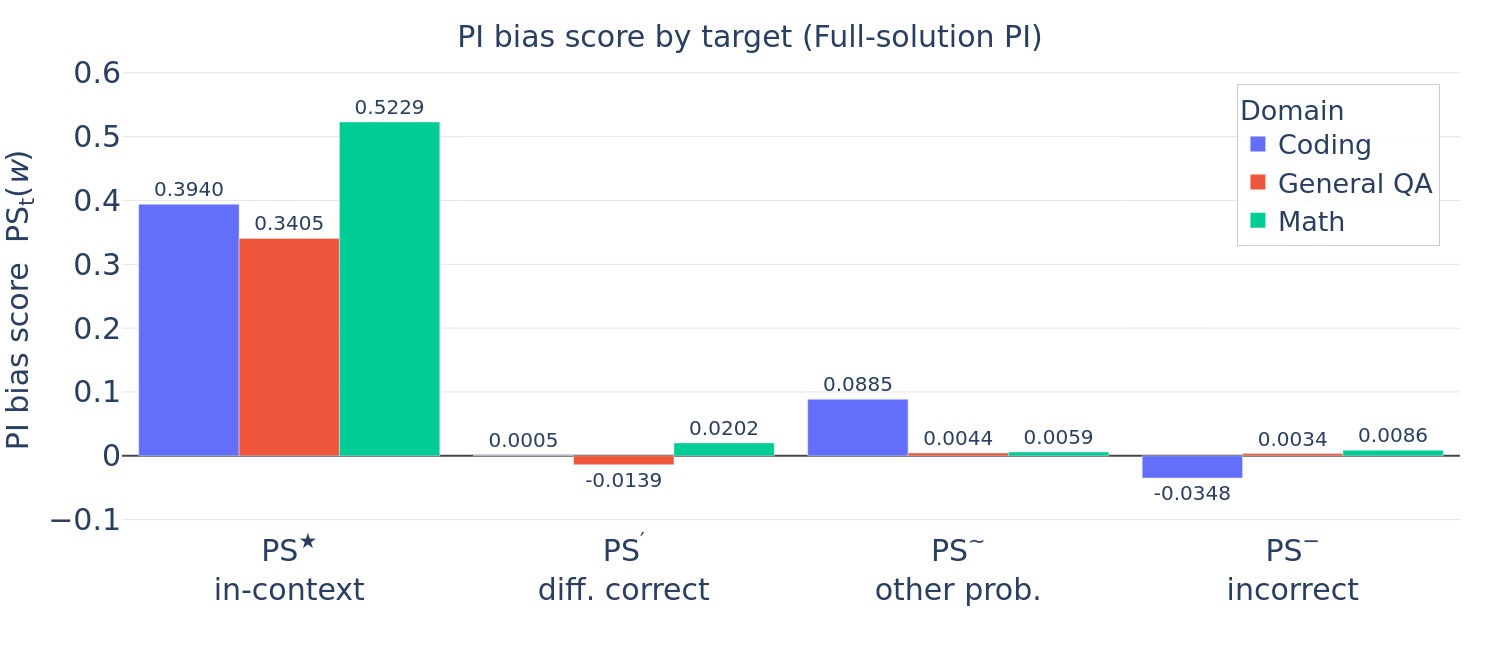}
\caption{\textbf{The teacher targets one solution, not correctness.} PI bias score (Eq.~\eqref{eq:pibias}) per domain, whole-solution PI; taller bar $=$ more mass moved toward that target. Only $\pibias^{\star}$, the in-context solution, is tall; a different \emph{correct} solution $\pibias'$ scores like an unrelated one $\pibias^{\sim}$, where a correctness-encoding teacher would give $\pibias^{\star}\approx\pibias'$.}
\label{fig:pibias}
\end{figure}

\paragraph{A design that separates correctness from imitation.}
We evaluate Eq.~\eqref{eq:pibias} for four targets: the in-context solution $\ysol$, a \emph{different correct} solution $\yalt$ to the same query, a solution $\yother$ to an unrelated problem, and an incorrect solution $\ywrong$, written $\pibias^{\star}$, $\pibias'$, $\pibias^{\sim}$, $\pibias^{-}$. The contrast $\pibias^{\star}$ vs.\ $\pibias'$ discriminates: a teacher encoding \emph{correctness} would give $\pibias^{\star}\approx\pibias'$, one encoding \emph{the trajectory it was shown} would spike on $\pibias^{\star}$ alone. We observe the second, in every domain (Figure~\ref{fig:pibias}): $\pibias^{\star}\gg\pibias'\approx\pibias^{\sim}>\pibias^{-}$, the teacher barely separating a different correct solution from one to an unrelated problem. \textbf{The teacher does not encode correctness; it encodes one solution.} This is the empirical counterpart of the bound of \citet{rlsd2025}, turning their irreducible $I(Y_t;R\mid X,Y_{<t})$ into a measurable quantity; the gap $\pibias^{\star}-\pibias'$ persists throughout training.

\paragraph{Weaker PI gives a weaker signal, not a better one.}
\label{sec:pichoice}
If over-specificity were the whole problem, a more general PI should repair it. On DAPO-Math we replace the whole solution with the two weaker forms of Section~\ref{sec:setup}, short \textbf{hints} and detailed \textbf{skills}. The bias moves as predicted, but not usefully: under both, the score collapses for every target, with $\pibias^{\star}$, $\pibias'$, $\pibias^{-}$ and $\pibias^{\sim}$ close together. For hints as PI, the PS* is 0.0015, PS- is 0.0192 and PS\~ is -0.0578, for skills as PI, the PS* is -0.0021, PS- is -0.0108 and PS\~ is 0.0071. Note that for hints and skills, PS* and PS' do not differ in definition. The teacher no longer singles out the in-context solution, but now moves little mass toward \emph{any} target. Performance follows (Table~\ref{tab:pichoice}): against a base average of $63.4$, hints lose $5.1$ points and skills $3.5$, while the whole solution stays within half a point of base. The two regimes fail for opposite reasons, which is the sense in which the bias is causal rather than incidental: PI specificity controls the bias and neither end of the dial produces learning, a whole solution tying the target to one trajectory while hints and skills are too weak to move the student. Examples of hints, skills and how we construct them are detailed in Appendix~\ref{app:qual}.

\definecolor{fullbg}{HTML}{E8F5E9}
\definecolor{hintbg}{HTML}{FFF8E1}
\definecolor{skillbg}{HTML}{E8F0FE}

\definecolor{fulltxt}{HTML}{2E7D32}
\definecolor{hinttxt}{HTML}{B26A00}
\definecolor{skilltxt}{HTML}{1565C0}

\begin{table}[tb]
\centering
\caption{\textbf{Neither over-specific nor under-specific PI produces learning.} SD on DAPO-Math (Qwen3-8B think) under three forms of privileged information; hints and skills reduce PI bias (Figure~\ref{fig:pibias}) but weaken the signal.}
\label{tab:pichoice}

\footnotesize
\setlength{\tabcolsep}{2.8pt}
\renewcommand{\arraystretch}{1.0}

\sdgreset{pibase}
\sdgreset{piF}
\sdgreset{piH}
\sdgreset{piK}

\resizebox{\columnwidth}{!}{%
\begin{tabular}{@{}lccccccc@{}}
\toprule

& &
\multicolumn{2}{c}{\textbf{Full}} &
\multicolumn{2}{c}{\textbf{Hint}} &
\multicolumn{2}{c}{\textbf{Skill}} \\

\cmidrule(lr){3-4}
\cmidrule(lr){5-6}
\cmidrule(lr){7-8}

\textbf{Benchmark} &
\textbf{Base} &
\textbf{SD} & $\Delta$ &
\textbf{SD} & $\Delta$ &
\textbf{SD} & $\Delta$ \\

\midrule

DAPO-Math      & \sdb{pibase}{79.1}[2.0] & \sds{piF}{80.1}[2.0] & \sds{piH}{75.8}[2.1] & \sds{piK}{76.3}[2.1] \\
\textit{AIME24}        & \sdb{pibase}{67.5}[7.7] & \sds{piF}{66.0}[7.8] & \sds{piH}{59.0}[7.9] & \sds{piK}{63.2}[8.0] \\
\textit{AIME25}        & \sdb{pibase}{53.3}[7.4] & \sds{piF}{52.9}[7.4] & \sds{piH}{47.5}[8.2] & \sds{piK}{49.9}[8.1] \\
\textit{OlympiadBench} & \sdb{pibase}{53.6}[1.9] & \sds{piF}{53.4}[1.8] & \sds{piH}{50.7}[1.8] & \sds{piK}{50.3}[1.8] \\

\midrule

\textbf{Average}
&
\sdbavg{pibase}
&
\sdsavg{piF}{pibase}
&
\sdsavg{piH}{pibase}
&
\sdsavg{piK}{pibase}
\\

\bottomrule
\end{tabular}
}
\end{table}

\begin{figure*}[b]
\begin{adjustwidth}{-\figover}{-\figover}
\centering
\begin{subfigure}[t]{0.26\linewidth}
    \centering
    \includegraphics[width=\linewidth]{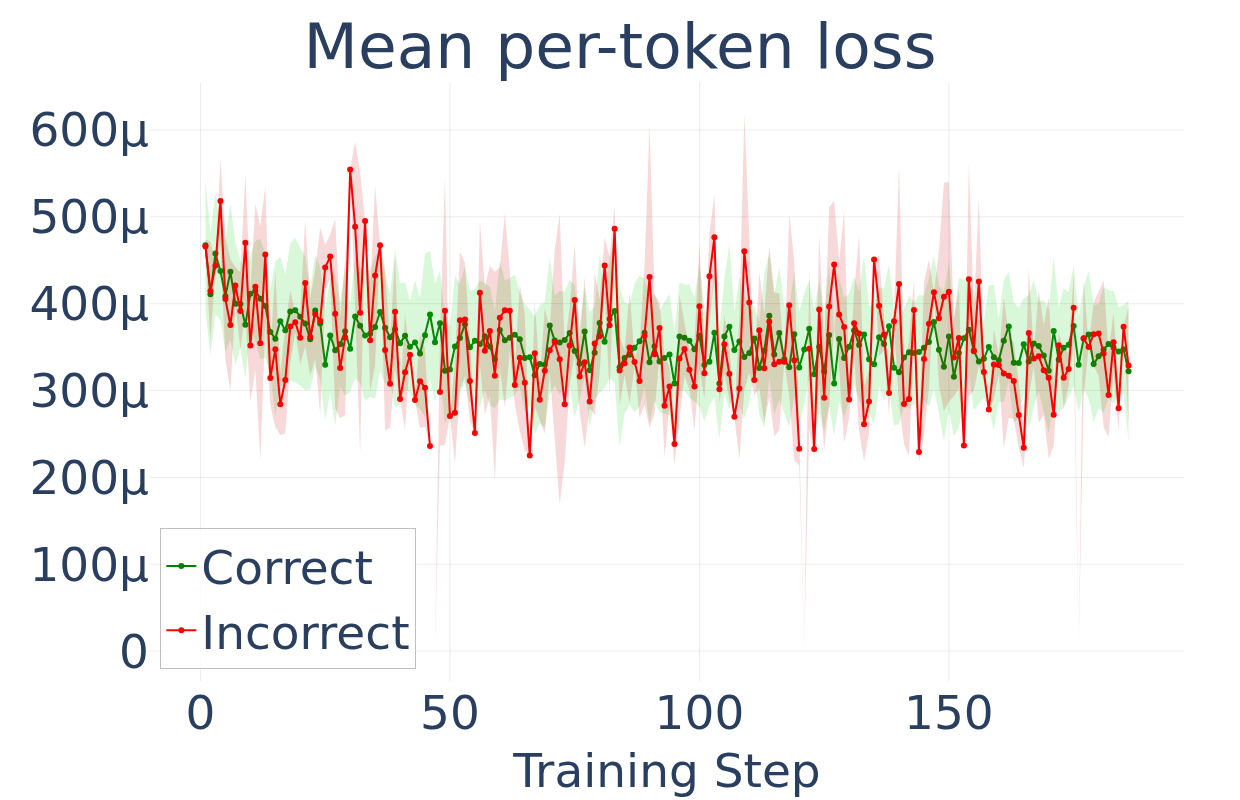}
    \caption{Mean per-token loss: curves overlap near $3.5\times10^{-4}$.}
    \label{fig:arg1-loss}
\end{subfigure}
\begin{subfigure}[t]{0.26\linewidth}
    \centering
    \includegraphics[width=\linewidth]{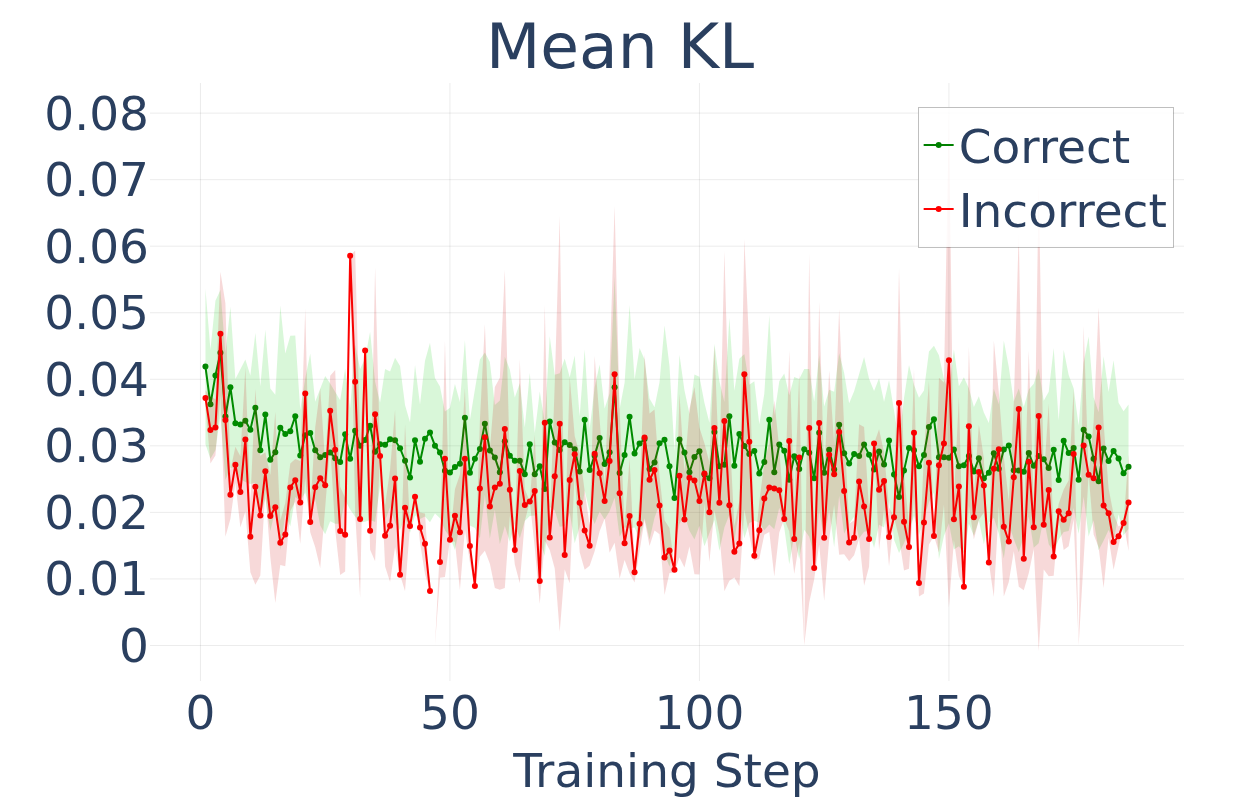}
    \caption{Mean KL: curves overlap near $0.03$.}
    \label{fig:arg1-kl}
\end{subfigure}
\begin{subfigure}[t]{0.26\linewidth}
    \centering
    \includegraphics[width=\linewidth]{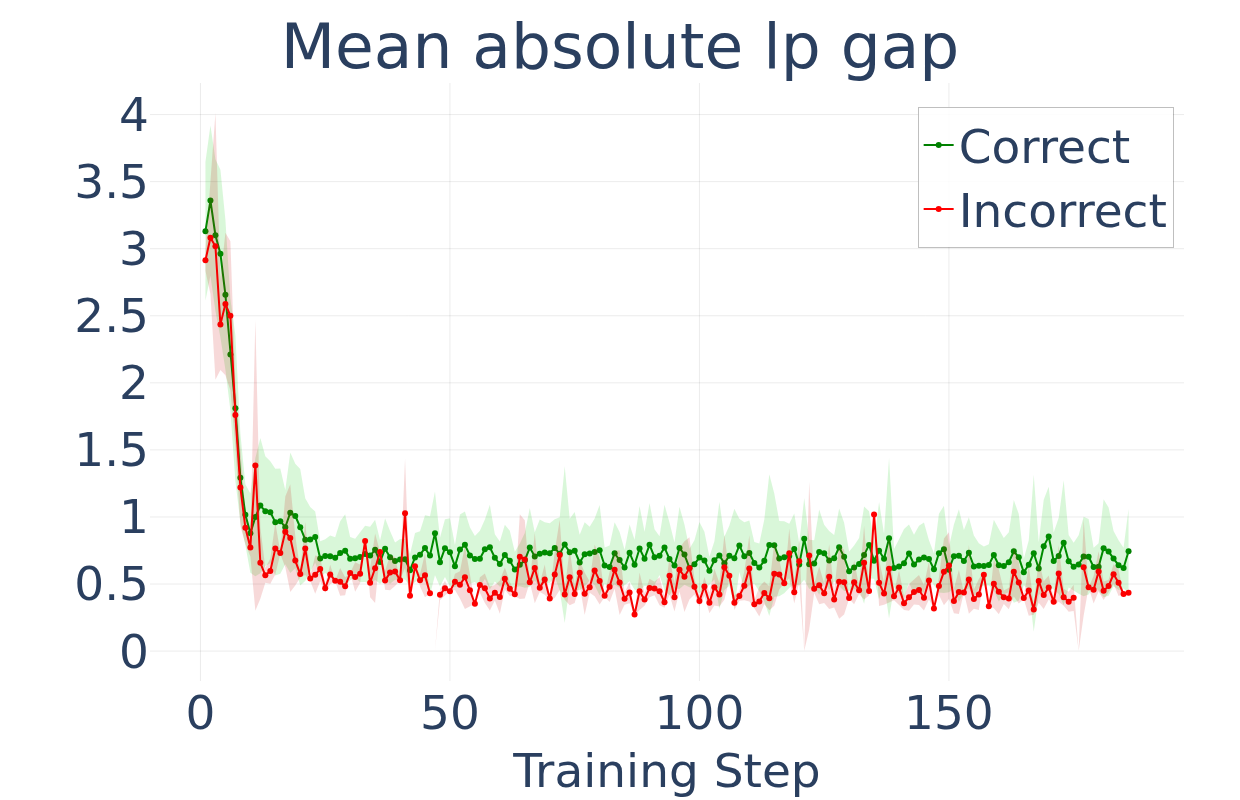}
    \caption{LP gap: correct ${\sim}0.7$, incorrect ${\sim}0.45$.}
    \label{fig:arg1-gap}
\end{subfigure}

\vspace{0.3em}

\begin{subfigure}[b]{0.3\linewidth}
\centering
\includegraphics[width=\linewidth]{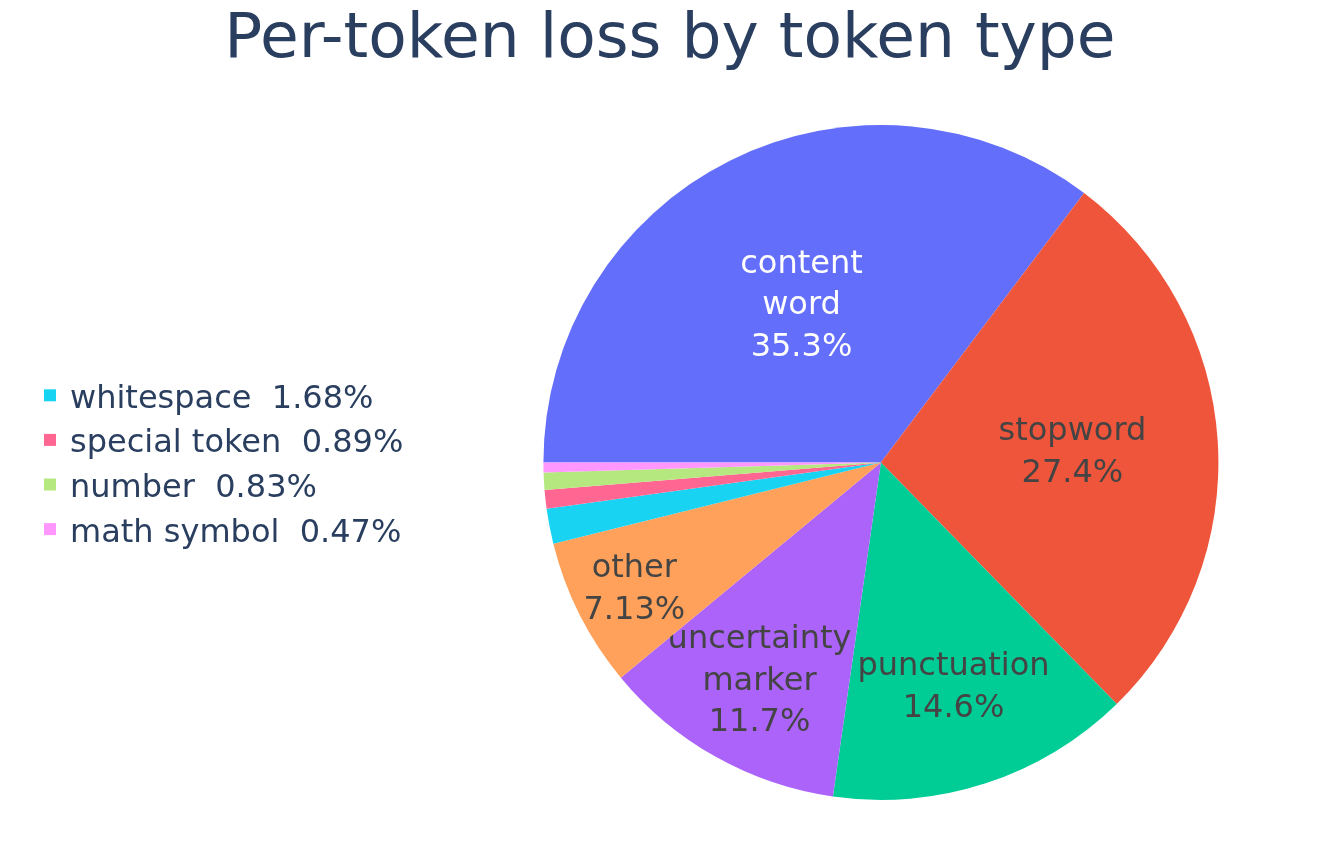}
\caption{Loss by token type}
\label{fig:arg2}
\end{subfigure}
\begin{subfigure}[b]{0.2\linewidth}
\centering
\includegraphics[width=\linewidth]{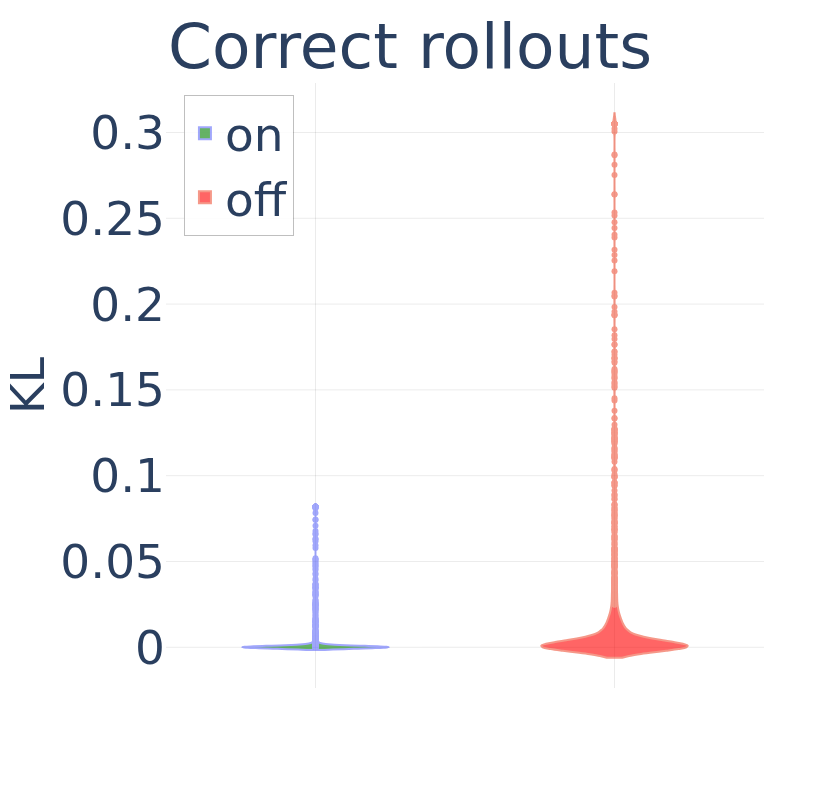}
\caption{Off- vs.\ on-path KL}
\label{fig:arg4}
\end{subfigure}
\begin{subfigure}[b]{0.3\linewidth}
\centering
\includegraphics[width=\linewidth]{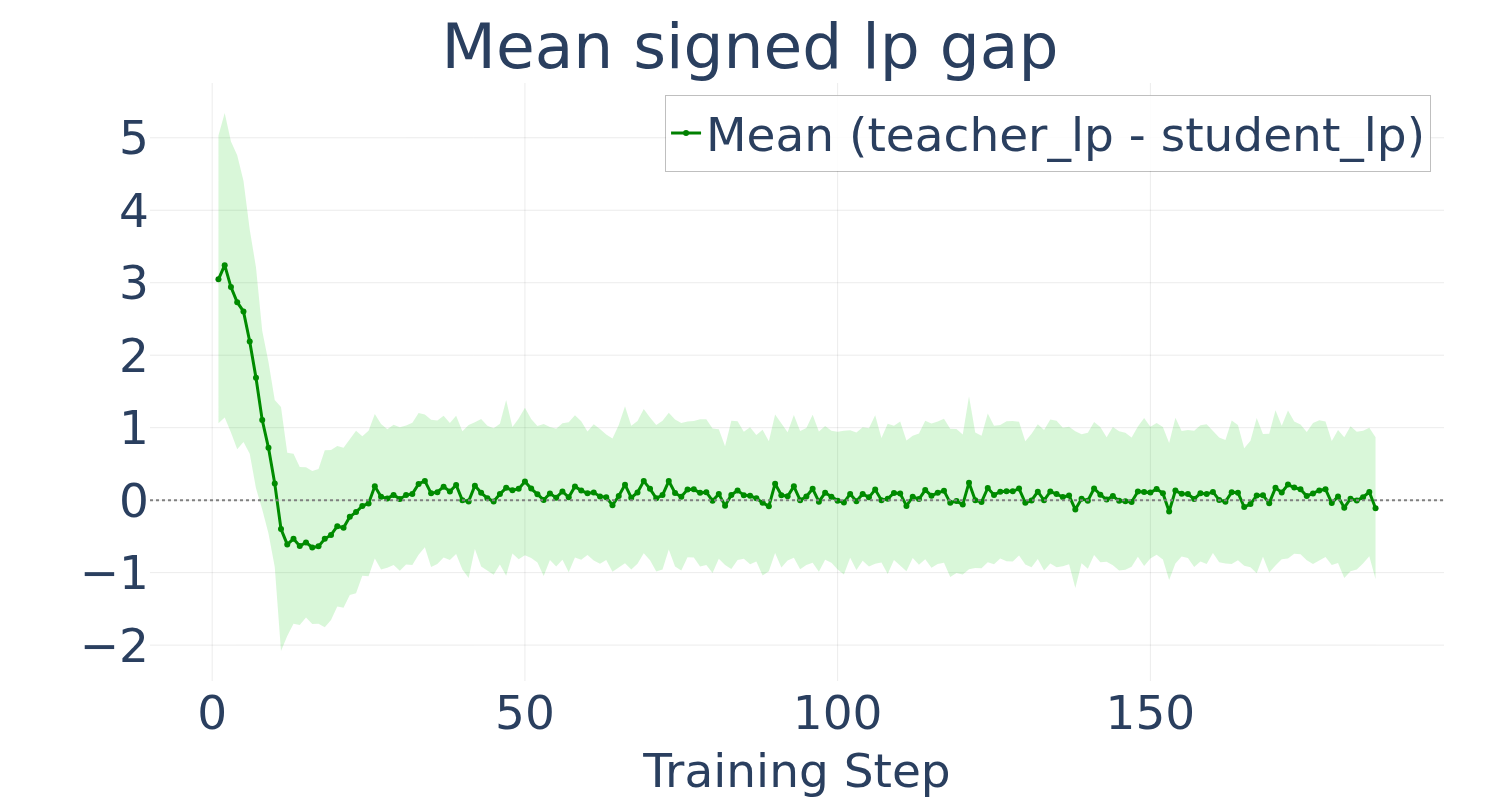}
\caption{Signed LP gap}
\label{fig:arg5}
\end{subfigure}

\end{adjustwidth}
\caption{\textbf{The causal chain.}
\textbf{(a--c) The loss does not track correctness}: for correct (green) vs.\ incorrect (red) rollouts, loss (a) and KL (b) overlap throughout training, while the absolute LP gap (c) separates them in the \emph{opposite} direction, staying higher on rollouts the student already solves.
\textbf{(d)} The loss budget falls mostly on low-information tokens rather than the content words, numbers and math symbols that determine the answer.
\textbf{(e)} Even within correct rollouts, off-path exploratory tokens incur much larger divergence than on-path ones.
\textbf{(f)} The average signed teacher--student gap shrinks to zero while its variance persists, indicating cancellation rather than convergence.}
\label{fig:chain}
\end{figure*}

\subsection{The loss does not track correctness}
\label{sec:arg1}
A target defined by one trajectory carries no information about whether the student's own trajectory reaches the correct answer, so Eq.~\ref{eq:sd} predicts a loss blind to correctness. An objective that taught problem-solving should assign greater loss to failures. It does not. Mean per-token loss settles near $3.5\times10^{-4}$ and mean KL near $0.03$ for both correct and incorrect rollouts, with overlapping four-rollout bands throughout training (Figures~\ref{fig:arg1-loss},~\ref{fig:arg1-kl}); if anything, incorrect rollouts exhibit slightly \emph{lower} KL. Thus, at no scale does the objective concentrate effort where the student goes wrong. Where the groups do separate, the direction is reversed: the \emph{absolute} log-probability gap $|\log p_T-\log p_S|$ falls from roughly $3$ to below $1$ within ten steps, then stabilizes near $0.7$ on correct rollouts versus $0.45$ on incorrect ones (Figure~\ref{fig:arg1-gap}). \textbf{The learning signal is strongest on rollouts the student already solves.} This directly explains the decoupling in Section~\ref{sec:signature}: a loss that cannot distinguish success from failure can decrease without improving accuracy.

\subsection{The loss lands on uninformative tokens}
\label{sec:arg2}
Per-token density predicts which tokens absorb the loss Section~\ref{sec:signature} showed to be concentrated. Labelling every token by one of nine mutually exclusive types (special token, whitespace, punctuation, uncertainty marker, stopword, number, math symbol, content word, or other) and attributing each step's loss accordingly, the tokens absorbing the most are predominantly low-information: in a representative step, stopwords, uncertainty markers (``wait,'' ``maybe,'' ``perhaps''), punctuation and whitespace account for \textbf{55.38\%} of the total per-token loss, while the content words, numbers and math symbols that determine the answer absorb comparatively little (Figure~\ref{fig:arg2}); the ordering is stable across training. Each token is assigned the first category it matches in the order listed above, with the uncertainty-marker set taken from \citet{opsd2025} and a standard English stopword list. \textbf{The objective's optimization capacity is spent primarily on tokens that do not determine correctness.} This unifies two prior observations: \citet{opsd2025} that stylistic tokens carry higher divergence than mathematical ones, \citet{reasoning_degrades2025} that SD shifts epistemic-marker probabilities by an order of magnitude more than the average token. Both are overloaded into one loss budget in which uninformative tokens dominate throughout training, not only at initialization.

\begin{figure}[tbp]
\centering
\includegraphics[width=\columnwidth]{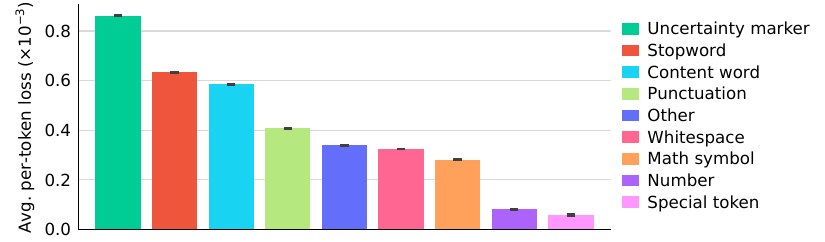}
\caption{Average per-token distillation loss for each token category. Bars are
ordered by decreasing average loss and whiskers denote $\pm 1$ SEM. Because the
loss is divided by the number of tokens in each category, a category cannot rank
highly merely by being numerous.}
\label{fig:toktypes}
\end{figure}

\paragraph{Average per-token loss by token type.}
Figure~\ref{fig:toktypes} reports the average per-token distillation loss for
each category, computed as the summed per-token loss of a category's tokens
divided by the number of those tokens. Dividing by the token count rather than
summing the loss mass means that a category cannot rank highly merely by being
numerous. Uncertainty markers carry the highest average per-token loss, followed
by stopwords and content words, while numbers and special tokens carry the
lowest. Because of this, even when the number of uncertainty markers and punctuations are less, their total loss share is high.

\subsection{The objective penalizes exploration}
\label{sec:arg4}
A reasoning model explores several partial paths before committing, so parts of even a correct rollout are legitimately off-path, and a teacher conditioned on $\ysol$ cannot recognize such a detour as productive. Ranking rollout positions by the student score $\sstud_t(\ysol)$, we call the top quartile \emph{on-path} and the bottom quartile \emph{off-path} and compare their full KL distributions, restricting first to correct rollouts so that off-path cannot be conflated with wrong. Within correct rollouts, off-path KL extends to ${\approx}0.31$ against ${\approx}0.08$ on-path, a factor of four (Figure~\ref{fig:arg4}). Whereas, within incorrect rollouts the asymmetry widens, ${\approx}0.47$ against ${\approx}0.10$. Because the teacher has read $\ysol$, a detour on the way to a correct answer is indistinguishable from a mistake, and these positions absorb the largest updates. \textbf{Within rollouts the student gets right, the objective penalizes exploration most, treating the search that reasoning requires as error.} This supplies the mechanism behind an effect \citet{reasoning_degrades2025} report only as a phenomenon, and it is why responses shorten: a model penalized for deliberating commits sooner.

\subsection{The student flattens rather than sharpens}
\label{sec:arg5}
The accumulated pressure leaves a trace on the student's distribution. The mean teacher--student log-probability gap falls from ${\approx}3.1$ at initialization, undershoots to ${\approx}-0.65$ around step~15, and settles at ${\approx}0$ from roughly step~30 (Figure~\ref{fig:arg5}), even with an EMA rate as small as $0.001$. Read alone, a gap of zero says the student has matched its teacher; two observations rule that out. The $\pm1$ standard-deviation band does not contract with the mean, staying near $\pm1$: the student matches the teacher \emph{on average} while disagreeing as much as ever at individual tokens, which indicates cancellation rather than convergence. And over the same period student entropy rises in most domains, with the teacher's perplexity on student rollouts rising alongside, whereas a student that had genuinely learned the teacher's distribution would become sharper and easier to predict. \textbf{The shrinking gap is not convergence but collapse: the student ends flatter and less decisive than it began}, closing the loop with Section~\ref{sec:signature}: a flatter policy commits earlier (falling length) and solves no more problems (flat validation).

\vspace{-5pt}
\section{Robustness}
\label{sec:further}

\textbf{Response length.}
\label{sec:lengthsplit}
A dense per-token loss spreads over more positions as trajectories lengthen, so the failure could be dilution; if it were, the short split would improve. Training separately on DAPO-Math response-length tertiles (short $[1318,4364]$, medium $[4387,11939]$, long $[11972,16384]$ tokens) all three fall below the base average of $63.4$, by $1.1$, $0.8$ and $1.6$ points; in-domain DAPO-Math rises in every split ($+1.4$, $+1.0$, $+0.6$) while eight of the nine transfer cells fall and the ninth gains $0.1$ in Table~\ref{tab:resp_len}. The short split, where supervision is least diluted, is no better than the long one, and its token-type loss shares are unchanged.

\begin{table}[t]
\centering
\caption{\textbf{Robustness to trajectory length.}
Response-length tertiles on DAPO-Math (short $[1318,4364]$, medium $[4387,11939]$, long $[11972,16384]$ tokens); Qwen3-8B think.
``SD'' and $\Delta=\text{SD}-\text{Base}$ follow Table~\ref{tab:main}, \textcolor{chgpos}{green} positive and \textcolor{chgneg}{red} negative; base scores are omitted for space and recoverable as $\text{SD}-\Delta$.}
\small
\renewcommand{\arraystretch}{1.1}
{\setlength{\tabcolsep}{7pt}\sdgreset{lenbase}\sdgreset{lenS}\sdgreset{lenM}\sdgreset{lenL}%
\begin{tabular}{@{}l*{6}{c}@{}}
\toprule
& \multicolumn{6}{c}{\cellcolor{blue!8}\textbf{Response Length (Qwen3-8B)}} \\
\cmidrule(lr){2-7}
& \multicolumn{2}{c}{Short} & \multicolumn{2}{c}{Medium} & \multicolumn{2}{c}{Long} \\
\cmidrule(lr){2-3}\cmidrule(lr){4-5}\cmidrule(lr){6-7}
\textbf{Benchmark} & SD & $\Delta$ & SD & $\Delta$ & SD & $\Delta$ \\
\midrule
DAPO-Math & \sdbq{lenbase}{79.1}[2.0]\sds{lenS}{80.5}[1.9] & \sds{lenM}{80.1}[1.9] & \sds{lenL}{79.7}[2.0] \\
AIME24 & \sdbq{lenbase}{67.5}[7.7]\sds{lenS}{62.9}[7.5] & \sds{lenM}{65.9}[7.5] & \sds{lenL}{63.8}[7.4] \\
AIME25 & \sdbq{lenbase}{53.3}[7.4]\sds{lenS}{52.7}[7.8] & \sds{lenM}{50.7}[7.5] & \sds{lenL}{50.6}[7.8] \\
Olympiad & \sdbq{lenbase}{53.6}[1.9]\sds{lenS}{53.0}[1.8] & \sds{lenM}{53.7}[1.8] & \sds{lenL}{53.1}[1.8] \\
\cmidrule(l){1-7}
Average & \sdsavg{lenS}{lenbase} & \sdsavg{lenM}{lenbase} & \sdsavg{lenL}{lenbase} \\
\bottomrule
\label{tab:resp_len}
\end{tabular}}
\end{table}

\textbf{Task difficulty.}
\label{sec:difficultysplit}
The failure could instead reflect problems the student cannot solve at all; if so, the easy split would improve. Splitting MMLU-Pro by the base model's mean success rate (MSR) at 8 samples into easy ($\mathrm{MSR@8}\in[0.75,0.875]$), medium ($[0.375,0.625]$) and hard ($[0.125,0.25]$) subsets: all three fall below the base average of $60.9$, easy by $3.6$ points, medium by $1.4$, hard by $1.6$ as seen in Table~\ref{tab:diff-scale}. In summary, the model performs no better on the easy split as compared to the hard split.

\textbf{Model scale.}
\label{sec:size}
SD's benefits are reported to grow with model size \citep{sdpo2025}, so a larger model should improve. Table~\ref{tab:diff-scale} shows how training Qwen3-32B on MMLU-Pro moves averages by $-0.6$ points in think and $-1.4$ in instruct, with the signature unchanged and the PI bias score undiminished. The larger model starts from a higher base, $64.7$ against $60.9$ for Qwen3-8B on the same benchmarks. With the reasoning modes of Section~\ref{sec:signature} and the PI-form intervention of Section~\ref{sec:arg3}, the result is therefore unchanged across trajectory length, task difficulty, model scale, reasoning mode, and PI content.

\begin{table}[tb]
\centering
\caption{\textbf{Robustness to task difficulty and model scale.}
Easy/medium/hard subsets of MMLU-Pro (left) and Qwen3-32B on full MMLU-Pro (right); Qwen3-8B think unless noted.
``SD'' and $\Delta=\text{SD}-\text{Base}$ follow Table~\ref{tab:main}, \textcolor{chgpos}{green} positive and \textcolor{chgneg}{red} negative; base scores are omitted for space and recoverable as $\text{SD}-\Delta$.}
\footnotesize
\renewcommand{\arraystretch}{1.0}
{\setlength{\tabcolsep}{2.4pt}\sdgreset{difbase}\sdgreset{difE}\sdgreset{difM}\sdgreset{difH}\sdgreset{sizeT}\sdgreset{sizeI}%
\resizebox{\linewidth}{!}{%
\begin{tabular}{@{}l*{10}{c}@{}}
\toprule

&
\multicolumn{6}{c}{\cellcolor{blue!8}\textbf{Difficulty Split (Qwen3-8B)}}&
\multicolumn{4}{c}{\cellcolor{green!8}\textbf{Model Scale (Qwen3-32B)}}\\

\cmidrule(lr){2-7}
\cmidrule(l){8-11}

&
\multicolumn{2}{c}{Easy}
&
\multicolumn{2}{c}{Medium}
&
\multicolumn{2}{c}{Hard}
&
\multicolumn{2}{c}{Think}
&
\multicolumn{2}{c}{Instruct}
\\

\cmidrule(lr){2-3}
\cmidrule(lr){4-5}
\cmidrule(lr){6-7}
\cmidrule(lr){8-9}
\cmidrule(l){10-11}

\textbf{Benchmark}
&
SD
&
$\Delta$
&
SD
&
$\Delta$
&
SD
&
$\Delta$
&
SD
&
$\Delta$
&
SD
&
$\Delta$
\\

\midrule

MMLU-Pro
&
\sdbq{difbase}{70.2}[2.2]
\sds{difE}{69.1}[2.2]
&
\sds{difM}{69.9}[2.2]
&
\sds{difH}{69.5}[2.3]
&
\sdgs{sizeT}{78.7}[1.9]{78.8}[2.0]
&
\sdgs{sizeI}{77.4}[2.1]{75.7}[2.1]
\\

GPQA-D
&
\sdbq{difbase}{59.5}[3.0]
\sds{difE}{58.3}[3.0]
&
\sds{difM}{58.7}[3.1]
&
\sds{difH}{57.4}[3.0]
&
\sdgs{sizeT}{68.4}[2.8]{66.5}[3.0]
&
\sdgs{sizeI}{52.4}[3.0]{49.9}[3.0]
\\

Bio (SKE)
&
\sdbq{difbase}{38.5}[3.4]
\sds{difE}{32.6}[3.8]
&
\sds{difM}{34.9}[3.8]
&
\sds{difH}{35.6}[3.8]
&
\sdgs{sizeT}{36.5}[3.6]{35.5}[3.4]
&
\sdgs{sizeI}{34.5}[3.3]{31.3}[3.3]
\\

Chem (SKE)
&
\sdbq{difbase}{51.9}[2.0]
\sds{difE}{51.8}[2.0]
&
\sds{difM}{51.9}[2.0]
&
\sds{difH}{52.2}[2.0]
&
\sdgs{sizeT}{59.9}[1.9]{57.9}[2.0]
&
\sdgs{sizeI}{50.4}[2.0]{50.9}[2.0]
\\

Mat. (SKE)
&
\sdbq{difbase}{68.0}[3.2]
\sds{difE}{64.7}[3.1]
&
\sds{difM}{64.6}[3.1]
&
\sds{difH}{64.5}[3.1]
&
\sdgs{sizeT}{68.3}[3.1]{68.4}[3.1]
&
\sdgs{sizeI}{68.3}[3.0]{68.7}[3.0]
\\

Phys. (SKE)
&
\sdbq{difbase}{77.5}[2.8]
\sds{difE}{67.5}[3.1]
&
\sds{difM}{77.0}[3.1]
&
\sds{difH}{76.9}[2.9]
&
\sdgs{sizeT}{76.3}[3.0]{77.3}[2.9]
&
\sdgs{sizeI}{69.8}[3.0]{68.0}[3.2]
\\

\cmidrule(l){1-11}

\textit{Average}
&
\sdsavg{difE}{difbase}
&
\sdsavg{difM}{difbase}
&
\sdsavg{difH}{difbase}
&
\sdgavgs{sizeT}
&
\sdgavgs{sizeI}
\\

\bottomrule
\label{tab:diff-scale}
\end{tabular}}}
\end{table}

\section{Discussion}
\label{sec:discussion}

\textbf{What fails is the target, not the density.} Not optimization, implementation, data, or scale (Sections~\ref{sec:signature},~\ref{sec:further}), but the target in Eq.~\eqref{eq:sd}: a PI-conditioned teacher predicts \textit{one} correct continuation and is indifferent between it and the others, so it spreads trajectory information, not correctness, over every token. Density cannot repair a target that cannot distinguish a correct continuation from a well-phrased incorrect one.

\textbf{Why the objective works on easy tasks.} Our account predicts rather than contradicts
prior work's regime of success: where responses are short and the answer is a bounded choice,
one reference solution nearly spans the space of correct ones, little exploration exists to
suppress, and few tokens are irrelevant to correctness, so imitating one solution (even at
high PI bias) is close to learning to be correct. All three weaken as responses lengthen and
answers open up; our results and SDPO/OPSD's are consistent,
differing only in where imitation/correctness cease to coincide.

\textbf{Implications for practice.} A falling SD loss is not evidence of
learning; monitor validation accuracy, which separates the two cases. The PI Bias Score
screens a candidate PI before training at a cost negligible beside it, forward passes over a
handful of prefixes and four targets: a large $\pibias^{\star}-\pibias'$ gap means the teacher
will transfer a trajectory rather than a skill, a uniformly small score that it will transfer
nothing. Nor is PI a free hyperparameter with a benign middle setting: both ends of the
specificity dial fail.

\textbf{Relation to methods that retain a reward, and the open problem.} The augmentation
literature \citep{rlsd2025,skillsd2025,sdar2025} divides by what each term supplies: a
verifiable reward carries correctness but nothing about form, the SD term form but almost
nothing about correctness. This is what RLSD's fix accomplishes: reweighting a verifier-driven
advantage keeps the density and discards the term carrying the bias. Whether combining them
repairs the failure we document, our experiments cannot answer: we do not run that setting. A
dense target that \textit{also} encodes correctness remains open; because the bias enters at
the first link, the teacher's commitment to a single trajectory, the constructions with the
clearest claim on the source make the target reflect the \textit{set} of correct continuations.

\textbf{Limitations.} Our study is observational, characterizing the objective rather than proposing a remedy. We evaluate one model family (Qwen3) at two scales across four domains and make no cross-architecture claim; results average four rollouts, and some per-domain differences are small relative to run-to-run variation. The PI Bias Score is a log-probability ratio at sampled positions, depends on the choice of $\yalt,\yother,\ywrong$, and complements the conditional mutual information analysis of \citet{rlsd2025} without estimating it. Hints and skills are model-generated, so the weak-PI results may partly reflect their construction. We do not claim self-distillation is ineffective in general, only that in our settings a PI-conditioned per-token objective alone gives no learning signal aligned with task correctness.
\vspace{-5pt}
\section{Conclusion}
\label{sec:conclusion}
We asked whether a dense PI-conditioned target carries information about correctness when it is the sole training signal. It does not. Across four reasoning-heavy domains, loss decreases while accuracy stays flat or degrades, under both recipes, both reasoning modes, and across trajectory length, task difficulty, and model scale. The PI Bias Score shows why: a teacher conditioned on a single reference solution learns that trajectory rather than correctness. The resulting loss cannot distinguish correct from incorrect rollouts, concentrates on already-correct trajectories, low-information tokens, and off-path exploration, and leaves the student flatter rather than sharper. Weakening the PI fails too, so neither overly specific nor overly weak PI signals correctness. The limitation is not target density but a teacher anchored to one reference trajectory rather than the task objective.



\clearpage
\bibliographystyle{plainnat}
\bibliography{references}

\clearpage

\appendix
\renewcommand{\thesection}{\Alph{section}}
\renewcommand{\thesubsection}{\Alph{section}.\arabic{subsection}}

\newcommand{\tocsec}[3]{\noindent\makebox[1.6em][l]{#1}#2\ \dotfill\ \pageref{#3}\par}
\newcommand{\tocsub}[3]{\noindent\hspace{1.6em}\makebox[2.2em][l]{#1}#2\ \dotfill\ \pageref{#3}\par\vspace{-0.15em}}
\section*{Contents}
{\small
\tocsec{A}{Datasets, Splits, and PI Construction}{app:data}
\tocsub{A.1}{Sources and splits}{app:sub:sources}
\tocsub{A.2}{PI construction}{app:sub:piconstruct}
\tocsec{B}{Hyperparameters (Implementation Details)}{app:hyper}
\tocsec{C}{Qualitative Examples (Full Solution vs Hints vs Skills)}{app:qual}
\tocsec{D}{Prompts and Templates}{app:prompts}
\tocsub{D.1}{Per-domain problem prompts}{app:sub:promptdomain}
\tocsub{D.2}{Self-distillation templates}{app:sub:sdtemplate}
\tocsub{D.3}{Privileged-information generation prompts}{app:sub:pigen}
\tocsub{D.4}{Skill derivation prompt}{app:sub:skill}
\tocsec{E}{PI Bias Implementation Details}{app:pibias}
}
\bigskip

This supplement provides implementation details, additional results, and
qualitative material for the main paper.

\clearpage

\section{Datasets, Splits, and PI Construction}
\label{app:data}

The study spans four domains: general question answering, mathematics, coding,
and multi-turn agentic tool use. For each domain, one dataset provides training
and in-domain evaluation, and a set of held-out datasets measures transfer.
Table~\ref{app:tab:domains} lists them. This section describes their sources, the
train, validation, and test splits, and the construction of the three forms of
privileged information (PI).

\begin{table}[t]
\centering
\footnotesize
\caption{Domains, training data, and evaluation benchmarks used in the study.}
\label{app:tab:domains}
\begin{tabular}{p{0.16\columnwidth}p{0.18\columnwidth}p{0.16\columnwidth}p{0.32\columnwidth}}
\toprule
\textbf{Domain} & \textbf{Train} & \textbf{In-domain} & \textbf{Transfer} \\
\midrule
General QA & MMLU-Pro (2k) & MMLU-Pro & GPQA-D, SciKnowEval \\
Mathematics & DAPO-Math (2k) & DAPO-Math & AIME24, AIME25, OlympiadBench \\
Coding & CodeForces (2k) & CodeForces & MBPP+, HumanEval+, CodeElo, LCBv6 \\
Multi-turn agentic & BFCLv3 Multi-turn base (100) & BFCLv3 Multi-turn base & BFCLv4 Multi-turn base, Miss.\ func, Miss.\ param, Long context \\
\bottomrule
\end{tabular}
\end{table}

\subsection{Sources and splits}
\label{app:sub:sources}

General question answering uses MMLU-Pro \citep{mmlupro2024} for training and
in-domain evaluation. Transfer is measured on GPQA-Diamond
\citep{gpqa2023} and on four SciKnowEval \citep{sciknoweval2024} subjects
(biology, chemistry, materials, and physics). The SciKnowEval subjects are
divided into training and test portions by a ninety-ten split.

Mathematics uses the processed release of DAPO-Math
\citep{dapo2025,dapomathprocessed2025} for training and in-domain evaluation, and
measures transfer on AIME~2024 \citep{aime2024}, AIME~2025 \citep{aime2025}, and
OlympiadBench \citep{olympiadbench2024}.

Coding uses CodeForces \citep{codeforcesdata2025} problems, restricted to those
with a problem description, for training and in-domain evaluation, and measures
transfer on MBPP+ and HumanEval+ \citep{mbpp2021,humaneval2021,evalplus2023},
CodeElo \citep{codeelo2025}, and LiveCodeBench~v6 \citep{lcb2025}.

The multi-turn agentic domain uses the Berkeley Function-Calling Leaderboard
\citep{bfcl2024}. Training and in-domain evaluation use the version-three
multi-turn base v3 category, which comprises 100 tasks. Transfer is measured on the
version-four multi-turn v4 categories: base, missing function, missing parameter,
and long context.

For the in-domain training datasets, a stratified validation subset of 300
examples is drawn, stratified over pass-rate bins. For the transfer datasets, the
validation set is the full dataset. At training time, each in-domain training set
is subsampled to a random subset of 2{,}000 examples drawn without replacement.
The multi-turn agentic domain is the exception and uses its full training set.

\subsection{PI construction}
\label{app:sub:piconstruct}

Three forms of privileged information are used. Any one of them can be supplied to
training in place of the others.

\paragraph{Whole solution.} For each question, one correct solution is selected
from the model's own successful attempts, with its reasoning trace removed. The
selected solution is presented to the teacher through the templates of
Appendix~\ref{app:prompts}.

\paragraph{Hint.} A short insight of one to two sentences is generated from the
question and a correct solution by a separate model, and is constrained not to
reveal the final answer. The generation prompt is given in
Appendix~\ref{app:prompts}.

\paragraph{Skill.} Problems are grouped into clusters, and one structured skill
card of roughly 500 words is derived per cluster and shared across its
member problems. Each card names a general technique together with its tools,
common pitfalls, and the cues that indicate when it applies, and is constrained
not to reveal any specific answer. The derivation prompt is given in
Appendix~\ref{app:prompts}.

\section{Hyperparameters (Implementation Details)}
\label{app:hyper}

We report the training configuration used for the self-distillation runs,
grouped following the format used in the appendix of the original SDPO paper
\citep{sdpo2025}. The configuration for the primary Qwen3-8B runs is shared across
the four domains; we note where the coding and multi-turn agentic domains and the
Qwen3-32B size study differ. The base model is Qwen3-8B \citep{qwen32025}, and
every domain is studied in both its thinking (long chain-of-thought) and instruct
variants; the configuration below corresponds to the thinking variant. All
experiments are run on a single node of eight NVIDIA B200 GPUs.

Two coefficients with similar names are distinguished here to avoid confusion.
The divergence used by the loss is controlled by a mixing coefficient for which
$0$ denotes forward KL, $1$ denotes reverse KL, and $0.5$ denotes the symmetric
Jensen-Shannon divergence; every primary run uses $0.5$. The rate at which the
teacher tracks the student through an exponential moving average is a separate
quantity, set to $0.001$; this is the teacher EMA rate referred to in the main
paper.

\begin{table}[t]
\centering
\footnotesize
\caption{Training configuration for the primary Qwen3-8B self-distillation runs,
grouped following the format used by the original SDPO paper \citep{sdpo2025}. The
configuration is shared across the four domains; the deviations noted below the
table apply to the coding, multi-turn agentic, and Qwen3-32B size-study runs.}
\label{app:tab:hyper}
\begin{tabular}{@{}p{0.62\columnwidth}p{0.30\columnwidth}@{}}
\toprule
\textbf{Parameter} & \textbf{Value} \\
\midrule
\multicolumn{2}{@{}l}{\textit{General}} \\
Model & Qwen3-8B \\
Thinking & enabled \\
\addlinespace
\multicolumn{2}{@{}l}{\textit{Data}} \\
Max.\ prompt length & 2048 \\
Max.\ response length & 16384 \\
\addlinespace
\multicolumn{2}{@{}l}{\textit{Batching}} \\
Question batch size & 32 \\
Mini batch size & 32 \\
Number of rollouts & 8 \\
\addlinespace
\multicolumn{2}{@{}l}{\textit{Rollout}} \\
Inference engine & vLLM \\
Temperature & 1.0 \\
\addlinespace
\multicolumn{2}{@{}l}{\textit{Validation}} \\
Number of rollouts & 4 \\
Temperature & 0.6 \\
Top-$p$ & 0.95 \\
\addlinespace
\multicolumn{2}{@{}l}{\textit{SDPO loss}} \\
Top-$K$ distillation & 100 \\
Distillation divergence & Jensen-Shannon ($\alpha = 0.5$) \\
Clip advantages & none \\
Teacher-EMA update rate & 0.001 \\
Rollout importance sampling clip & 2 \\
\addlinespace
\multicolumn{2}{@{}l}{\textit{Training}} \\
Optimizer & AdamW \\
Learning rate & $1\times10^{-5}$ (constant) \\
Warmup steps & 10 \\
Weight decay & 0.01 \\
Gradient clip norm & 1.0 \\
\addlinespace
\multicolumn{2}{@{}l}{\textit{Additional settings}} \\
Per-token loss clip $\tau$ & 0.001 \\
Max.\ reprompt length & 16384 \\
Total epochs & 3 \\
Training questions per epoch & 2000 \\
Validation frequency & every 5 steps \\
\bottomrule
\end{tabular}
\end{table}

The following domain-specific deviations from Table~\ref{app:tab:hyper} apply. The coding and multi-turn agentic runs use a learning
rate of $1\times10^{-6}$. The Qwen3-32B size study uses a maximum response length
of 10240, a maximum model context of 12288, and tensor-parallel size 4.

\section{Qualitative Examples (Full Solution vs Hints vs Skills)}
\label{app:qual}

\begin{figure*}[t]
\centering
\includegraphics[width=\textwidth]{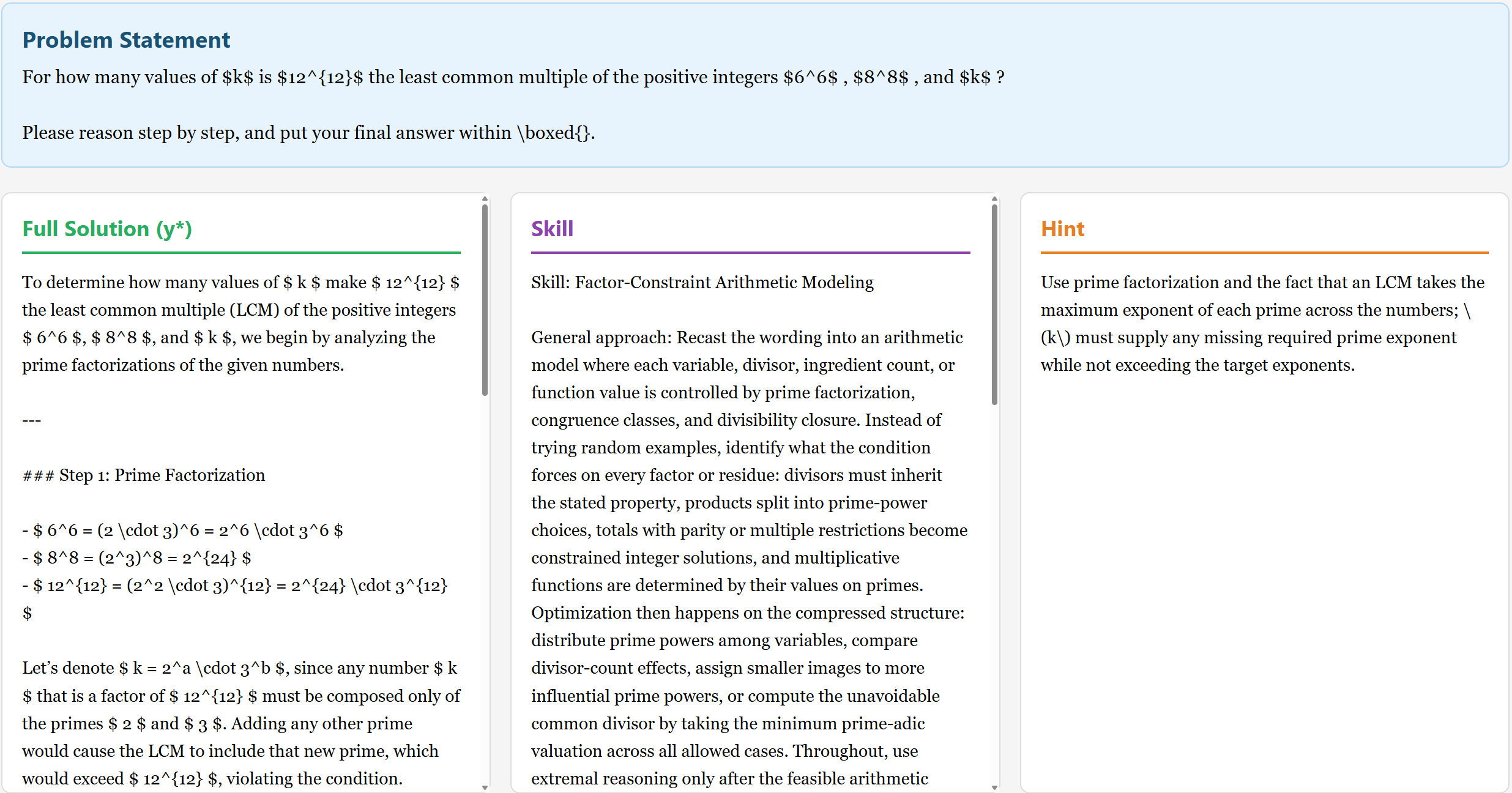}
\caption{A representative mathematics training question (top) shown with the three
forms of privileged information derived for it. The whole solution presents a
complete correct answer, the skill presents a structured technique card for the
question's cluster, and the hint presents a short insight. In training, each form
is supplied on its own to the privileged-information-conditioned teacher, following
the templates of Appendix~\ref{app:prompts}.}
\label{app:fig:qual}
\end{figure*}

This section illustrates the three forms of privileged information on a shared
training question. Figure~\ref{app:fig:qual} shows a representative mathematics
problem together with the whole solution, the derived skill, and the derived hint,
each of which can serve as the privileged information supplied to the teacher in a
separate training configuration.

\paragraph{Whole-solution form.} A single correct solution to the question, with
its reasoning trace removed, is presented to the teacher under a header that
labels it as the correct solution, followed by the instruction to solve the
original question with the student's own approach, following the mathematics
reprompt template of Appendix~\ref{app:prompts}.

\paragraph{Hint form.} A one-to-two-sentence insight is presented to the teacher
through the hint template of Appendix~\ref{app:prompts}. The insight names the
governing idea, for example the relevant theorem or identity, without stating the
final answer.

\paragraph{Skill form.} A structured skill card for the question's cluster is
presented to the teacher as five fields. The layout is
\begin{plst}
\begin{Verbatim}
Skill: <name>
General approach: <prose description of the technique>
Key tools & formulas: <theorems, identities, algorithms>
Pitfalls & checks: <common mistakes, edge cases, checks>
When it applies: <signals indicating this skill>
\end{Verbatim}
\listingcaption{Rendered skill-card fields.}
\end{plst}
The card describes a general technique for the cluster and does not reveal the
answer to any single problem.

\section{Prompts and Templates}
\label{app:prompts}

This section reproduces the prompt and template strings used in training and in
the construction of privileged information. Long lines are wrapped for display.
Placeholders in braces, for example \verb|{problem}| and \verb|{hint}|, are
filled at run time. Each block is labelled with a numbered listing caption.

\subsection{Per-domain problem prompts}
\label{app:sub:promptdomain}

The student is shown a domain-specific prompt that wraps each problem.

\begin{plst}
\begin{Verbatim}
{problem}

Please reason step by step, and put your final answer within \boxed{}.
\end{Verbatim}
\listingcaption{General reasoning and mathematics problem prompt.}
\end{plst}

\begin{plst}
\begin{Verbatim}
You are a coding expert. You will be given a coding problem, and you need to write a correct Python program that matches the specification and passes all tests. The time limit is 1 second. You may start by outlining your thought process. In the end, please provide the complete code in a code block enclosed with ``` ```.

{problem}
\end{Verbatim}
\listingcaption{Coding problem prompt.}
\end{plst}

\begin{plst}
\begin{Verbatim}
The following are multiple choice questions (with answers) about {subject}. Think step by step and then finish your answer with "the answer is (X)" where X is the correct letter choice.
Question:
{question}
Options:
A. {option_A}
B. {option_B}
...
Answer: Let's think step by step.
\end{Verbatim}
\listingcaption{MMLU-Pro multiple-choice prompt.}
\end{plst}

\begin{plst}
\begin{Verbatim}
Return your final response within \boxed{} and only include the letter choice (A, B, C, or D) as your final response.
Problem: {problem}
Options: {options}
Answer:
\end{Verbatim}
\listingcaption{GPQA prompt.}
\end{plst}

\begin{plst}
\begin{Verbatim}
Given a question and four options, please select the right answer. Respond in the following format:
<reasoning>
...
</reasoning>
<answer>
...
</answer>

For the answer, only output the letter corresponding to the correct option (A, B, C, or D), and nothing else. Do not restate the answer text. For example, if the answer is "A", just output:
<answer>
A
</answer>
\end{Verbatim}
\listingcaption{SciKnowEval system prompt.}
\end{plst}

\subsection{Self-distillation templates}
\label{app:sub:sdtemplate}

During self-distillation the teacher prompt is assembled from a reprompt template
that combines the problem, a solution section carrying the privileged information,
and an optional feedback section. The privileged information enters through the
solution section (a whole solution) or through the hint template (a hint or
skill). The default templates are shown first, followed by the per-domain
overrides used for mathematics and the multi-turn agentic setting; the overrides
label the injected privileged information as a correct solution and close the
reprompt with an instruction to use the student's own approach.

\begin{plst}
\begin{Verbatim}
{prompt}{solution}{feedback}

Correctly solve the original question.
\end{Verbatim}
\listingcaption{Default reprompt template.}
\end{plst}

\begin{plst}
\begin{Verbatim}
Correct solution:

{successful_previous_attempt}
\end{Verbatim}
\listingcaption{Default solution section.}
\end{plst}

\begin{plst}
\begin{Verbatim}
The following is feedback from your unsuccessful earlier attempt:

{feedback_raw}
\end{Verbatim}
\listingcaption{Default feedback section.}
\end{plst}

\begin{plst}
\begin{Verbatim}
Here is a hint for solving this problem:
{hint}
\end{Verbatim}
\listingcaption{Default hint template.}
\end{plst}

\begin{plst}
\begin{Verbatim}
hint_template:
  Here is the correct solution for the above question:

  {hint}
reprompt_template:
  {prompt}{solution}{feedback}

  Now, using your own approach, try to correctly solve the original question.
\end{Verbatim}
\listingcaption{Mathematics reprompt and hint templates.}
\end{plst}

\begin{plst}
\begin{Verbatim}
reprompt_template:
  {prompt}{solution}{feedback}

  After understanding this, please try to solve this problem using your own approach below.
solution_template:
  Here is the correct solution for the problem:
  ```
  {successful_previous_attempt}
  ```
\end{Verbatim}
\listingcaption{Multi-turn agentic reprompt and solution templates.}
\end{plst}

\subsection{Privileged-information generation prompts}
\label{app:sub:pigen}

Hints and compressed solutions used as privileged information are generated by a
separate model.

\begin{plst}
\begin{Verbatim}
You are given a math problem and its correct solution. Your job is to extract the core mathematical trick, principle, or insight that unlocks the solution.

Rules:
- 1-3 sentences maximum.
- Do NOT reveal the final answer or any numerical result.
- Do NOT give step-by-step instructions.
- Do NOT start with "The key insight is" or similar preambles.
- Only state the idea, theorem, formula, or trick itself.

Respond in the following format:
<hint>
[your hint here]
</hint>

Problem: ```{problem}```

Solution:
```
{solution}
```
\end{Verbatim}
\listingcaption{Hint generation prompt.}
\end{plst}

\begin{plst}
\begin{Verbatim}
You are given a math problem and its correct (but verbose) solution. Rewrite the solution as a short, dense derivation.

Rules:
- Include ALL key steps and the final answer.
- Remove all verbose explanations, motivations, headers, and formatting.
- Use mathematical notation directly instead of prose where possible.
- Target length: 3-6 sentences or equivalent equations.
- The compressed solution must be self-contained and verifiable.
- Do NOT omit the final boxed answer.

Respond in the following format:
<compressed>
[your compressed solution here]
</compressed>

Problem: ```{problem}```

Solution:
```
{solution}
```
\end{Verbatim}
\listingcaption{Compressed-solution generation prompt.}
\end{plst}

\subsection{Skill derivation prompt}
\label{app:sub:skill}

The skill form of privileged information is derived once per problem cluster.

\begin{plst}
\begin{Verbatim}
You are analyzing a cluster of related {domain} to extract ONE reusable problem-solving *skill* that applies across the whole cluster.

A skill is the Goldilocks middle ground between a one-line hint and a full solution: rich enough to materially guide a solver on ANY problem in this class, yet general enough that it never solves a specific instance. It will be shown to a model as privileged context alongside a long (multi-thousand-token) solution attempt, so it must carry enough substance to actually shape the reasoning.

Hard rules:
- Describe a GENERAL, reusable technique/strategy that plausibly applies to MANY problems in this cluster, not just one.
- Do NOT reveal or compute any specific numeric answer, final result, or option.
- Do NOT walk through the solution of any single problem instance.
- Do NOT give a numbered step-by-step procedure or a fixed solution template.
- Be concrete and substantive: name the actual reasoning moves, representations, formulas, theorems, and checks a strong solver would use.

Produce a structured skill card with these fields:
- name: a short 3-7 word handle for the technique.
- approach: the general line of attack, in flowing prose.
- tools: the key formulas, theorems, identities, representations, or algorithms this technique relies on.
- pitfalls: common mistakes, tricky edge cases, and sanity-checks.
- cues: the textual or structural signals in a problem that indicate this skill applies.

Aim for a substantial, information-dense card (roughly 250-400 words total across the fields).
\end{Verbatim}
\listingcaption{Skill derivation prompt.}
\end{plst}

\section{PI Bias Implementation Details}
\label{app:pibias}

This section documents how the PI Bias Score and the on-path score used in the
main paper are computed. For a student rollout, let the student prefix up to
position $t$ be its first $t$ response tokens. For a target token sequence $w$,
define the on-path score
\begin{equation}
s_t(w) = \frac{1}{|w|}\sum_{k=0}^{|w|-1}
\log P_{\text{model}}\!\left(w_k \mid c,\, \text{prefix}_{<t},\, w_{<k}\right),
\end{equation}
the mean log-probability the model assigns to the tokens of $w$ when $w$ is
appended after the student prefix. The context $c$ is either the teacher prompt,
which contains the privileged information, giving $\steach_t(w)$, or the student
prompt, which does not, giving $\sstud_t(w)$. The PI Bias Score is the difference
\begin{equation}
\pibias_t(w) = \steach_t(w) - \sstud_t(w),
\end{equation}
so that a positive value indicates that the privileged information present in the
teacher prompt raises the model's tendency to continue toward $w$.

\paragraph{Targets.}
Four targets $w$ are scored at each evaluated position:
\begin{itemize}
\item $\ysol$, the in-context correct solution, that is the privileged
  information itself, taken from the teacher prompt. For hint-style privileged
  information that does not embed a full solution, a correct solution to the same
  question is used with its reasoning trace removed.
\item $\yalt$, a different correct solution to the same question, which requires
  at least two distinct correct solutions.
\item $\ywrong$, an incorrect solution to the same question, taken from an
  incorrect attempt.
\item $\yother$, an unrelated correct solution drawn from the other questions in
  the same training step.
\end{itemize}

\paragraph{Positions and lookahead.}
Each target is scored in full: the lookahead length is set to $K = 2000$ tokens,
which exceeds the length of every target, so no target is truncated. Positions
are evaluated at every token, over assistant tokens (and not user tokens) only. Prompts longer than
8000 tokens are skipped. When the student is inside a reasoning block at position
$t$, a closing marker is inserted before the target so that the appended
continuation is well formed.

\paragraph{Models.}
The teacher score $\steach$ is computed with the teacher checkpoint and the
student score $\sstud$ with the student checkpoint at the corresponding training
step. Before the first checkpoint, the base model is used for both roles. The two
passes are computed separately and merged by position.

\paragraph{Scoring.}
The score is obtained from a single forward pass over the assembled prompt and
target. At each target position the language-model head produces a distribution
over the vocabulary, from which the log-probability of the target token is read,
and the on-path score is the mean of these log-probabilities over the target.

\end{document}